\documentclass[lettersize,journal]{IEEEtran}
\usepackage{amsmath,amsfonts}
\usepackage{algorithmic}
\usepackage{algorithm}
\usepackage{array}
\usepackage[caption=false,font=normalsize,labelfont=sf,textfont=sf]{subfig}
\usepackage{textcomp}
\usepackage{stfloats}
\usepackage{url}
\usepackage{verbatim}
\usepackage{graphicx}
\usepackage{cite}
\usepackage[numbers,sort&compress]{natbib}

\usepackage[normalem]{ulem}
\useunder{\uline}{\ul}{}

\begin{document}
\pdfoutput=1

\title{A Survey on Learnable Evolutionary Algorithms for \\ Scalable Multiobjective Optimization}

\author{Songbai Liu,~\IEEEmembership{Member,~IEEE,} Qiuzhen Lin,~\IEEEmembership{Member,~IEEE,} Jianqiang Li,~\IEEEmembership{Member,~IEEE,} \\ and Kay Chen Tan,~\IEEEmembership{Fellow,~IEEE}
        % <-this % stops a space
\thanks{Manuscript received August xx, 2022; revised October 2022, and January 2023; accepted February 2023. This work is partially supported by the National Natural Science Foundation of China (NSFC) under No. U21A20512, and in part by the Research Grants Council of the Hong Kong SAR under Grant PolyU11211521 and Grant PolyU15218622, and in part by The Hong Kong Polytechnic University (Project No.: 1-ZE0C), and in part by Shenzhen Science and Technology Program under Grants JCYJ20220531101411027, JCYJ20190808164211203 and R2020A045, and in part by Natural Science Foundation of Guangdong Province under Grant No. 2023A1515011238. (\textit{Corresponding authors: Qiuzhen Lin and Kay Chen Tan})}% <-this % stops a space
\thanks{S.B. Liu, Q.Z. Lin, and J.Q. Li are with the College of Computer Science and Software Engineering, Shenzhen University, Shenzhen 518060, China (e-mail: songbai@szu.edu.cn).}
\thanks{K. C. Tan is with the Department of Computing, The Hong Kong Polytechnic University, Hong Kong SAR (e-mail: kctan@polyu.edu.hk).}}

\markboth{Journal of IEEE Transactions on Evolutionary Computation,~Vol.~xx, No.~xx, February~2023}%
{Shell \MakeLowercase{\textit{et al.}}: A Survey on Learnable Evolutionary Algorithms for Scalable Multiobjective Optimization}

\IEEEpubid{0000--0000/00\$00.00~\copyright~2023 IEEE}

\maketitle

\begin{abstract}
Recent decades have witnessed great advancements in multiobjective evolutionary algorithms (MOEAs) for multiobjective optimization problems (MOPs). However, these progressively improved MOEAs have not necessarily been equipped with scalable and learnable problem-solving strategies for new and grand challenges brought by the scaling-up MOPs with continuously increasing complexity from diverse aspects, mainly including expensive cost of function evaluations, many objectives, large-scale search space, time-varying environments, and multi-task. Under different scenarios, divergent thinking is required in designing new powerful MOEAs for solving them effectively. In this context, research studies on learnable MOEAs with machine learning techniques have received extensive attention in the field of evolutionary computation. This paper begins with a general taxonomy of scaling-up MOPs and learnable MOEAs, followed by an analysis of the challenges that these MOPs pose to traditional MOEAs. Then, we synthetically overview recent advances of learnable MOEAs in solving various scaling-up MOPs, focusing primarily on four attractive directions (i.e., learnable evolutionary discriminators for environmental selection, learnable evolutionary generators for reproduction, learnable evolutionary evaluators for function evaluations, and learnable evolutionary transfer modules for sharing or reusing optimization experience). The insight of learnable MOEAs is offered to readers as a reference to the general track of the efforts in this field.
\end{abstract}


\section{Introduction}
\IEEEPARstart{E}{vol}utionary algorithms (EAs) are generally regarded as nondeterministic problem solvers inspired by biological evolution mechanisms such as natural selection and genetic variation. They can find a set of approximately optimal solutions for the problem without requiring domain-specific information such as continuous, differentiable, and unimodal functions. Using heuristic random search mechanisms, EAs have a strong ability to cope with uncertainty, which are suitable for finding global and robust optimal solutions \cite{jin2005uncertainEC}. In addition, EAs with population-based search are inherently capable of parallel computing \cite{alba2002parallEC}. More importantly, most EAs can obtain a set of solutions in a single run, so they have advantages in solving multiobjective optimization problems (MOPs) compared with mathematical programming approaches \cite{coello2020EMO}. Thus, EAs are widely used in many real-world applications to solve specific MOPs that have plagued traditional mathematical optimizers \cite{arias2012EMOAPP1, ponsich2013EMOAPP2}. Especially, multiobjective EAs (MOEAs) have been applied to tackle MOPs in other research areas, e.g., evolutionary machine learning (ML) \cite{zhan2022ECML, jin2006MOML}, evolutionary data mining \cite{telikani2022EML, mukhopadhyay2014ECDM}, evolutionary automatic network design \cite{zhou2021ECDNN}, and evolutionary transfer learning \cite{nguyen2021ECTL}.
\IEEEpubidadjcol
This progress in various areas, along with the increasing complexities faced by engineering and science, suggests that MOEAs are effective for problem-solving in many applications.

In general, an MOEA starts with an initial parent population of randomly sampled solutions. Then, it reproduces a population of offspring solutions (with some features inherited from their parents) by a generative or reproductive model. This model, termed evolutionary generator, is mainly equipped with some genetic operators \cite{bonissone2006evolutionary}, including mating selection, crossover, mutation, etc. All newly generated solutions need to go through a function evaluator to calculate the value of each of their objectives. After that, the combined population of parent and offspring is inputted into a discriminative (or selective) model. This model, referred to as evolutionary discriminator, outputs only the elite solutions discriminated to be fitter for surviving in the next generation (i.e., the natural or environmental selection based on the ``survival of the fittest'' principle \cite{salomon1998evolutionary}). From the perspective of ML as shown in Fig. 1, an MOEA's generator aims to reproduce offspring with superior qualities over their parents by evolution, while its discriminator aims to distinguish the qualities between parent and offspring solutions after the evaluator, followed by selecting a certain number of elites to the next generation. Together, they form the driving force for searching diverse and progressively converged solutions to approximate the unknown optima of the given MOPs \cite{drugan2019reinforcement}.

After decades of development, MOEAs should have grown into sophisticated, innovative, and creative solvers for various MOPs, especially from the perspective of biological evolution \cite{miikkulainen2021biological}. However, most existing MOEAs often lack the flexibility, plasticity, and scalability to effectively respond to major challenges brought by scaling-up MOPs. These problems grow in complexity as they are scaled up from diverse aspects, mainly including the expensive cost of function evaluations \cite{santana2010review}, many objectives \cite{li2015many}, large-scale decision variables \cite{omidvar2021review}, time-varying environments \cite{goh2007investigation}, and multitasking \cite{wei2021review}. The potential obstacles that degrade the performance of traditional MOEAs in solving these scaling-up MOPs are elaborately summarized in Section II.B.
\begin{figure}[!t]
\centering
\includegraphics[width=2.5in]{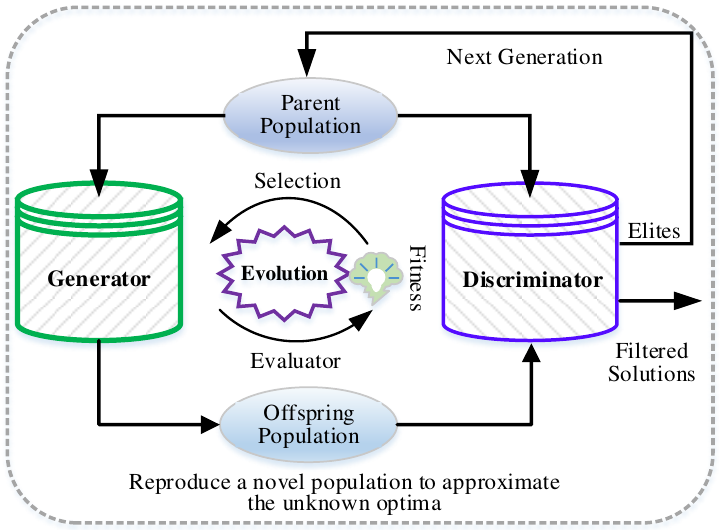}
\caption{The general flow of an MOEA from the perspective of ML.}
\label{fig_1}
\end{figure}
Moreover, with the rapid increase in the scale and complexity of today's engineering designs and available data, existing problem-solving systems cannot comprehend the ``big dimensionality'' spaces \cite{zhai2014emerging} or adaptively cope with all potential challenges brought by ``big complex optimization problems'' \cite{zhou2014big}. It is becoming pressingly demanded to design more intelligent and powerful MOEAs for tackling increasingly complex MOPs. Evolutionary optimization and incremental learning are two instinctual ways for humans to improve their problem-solving abilities. Thus, solving the MOPs only by evolution in traditional MOEAs may be inadequate and inefficient, especially when solving those scaling-up MOPs with diverse complexities. As stated in \cite{miikkulainen2021biological}, biological evolution may take thousands of years to optimize a species, while cumulative learning can help speed up the optimization process dramatically. Inspired by this, an insightful research question was born, i.e., can ML be incorporated into MOEAs (i.e., learnable MOEAs) to form a more powerful optimizer that solves scaling-up MOPs efficiently?

In this context, research studies on learnable MOEAs have received extensive attention in the field of evolutionary computation \cite{zhang2011evolutionary}. ML techniques are applied to assist and catalyze the evolutionary modules (i.e., generator, discriminator, and evaluator) in learnable MOEAs. Particularly, the generator of an MOEA can generate a huge volume of data (feasible solutions) by iteratively searching in the variable space. Systematic analysis of these data by ML techniques is helpful to better understand the search behavior and improve its future search capability \cite{goldberg1988genetic}. Exploring along promising directions learned in the search space, the MOEA's generator can effectively find solutions with great potential \cite{lin2018hybrid}. Through online predicting Pareto Front (PF) shapes, the MOEA's discriminator will be more competent to adaptively filter out poorly performing solutions when solving MOPs with various irregular PFs \cite{liefooghe2020landscape}. The objective and search spaces can be simplified with the help of dimension reduction and spatial transformation techniques before being tackled \cite{li2022offline}. Reinforcement learning (RL) can be employed to determine appropriate evolutionary operators (i.e., actions) given any parent (i.e., state), which can guide the generator to reproduce high-quality offspring \cite{jiao2023reference}. Domain adaptation can be used to learn the domain-invariant feature representations between different MOPs and then analyze their distribution divergence, which guides the knowledge transfer when solving them sequentially or simultaneously \cite{tan2021evolutionary}.

Therefore, during the evolutionary process, MOEAs incessantly produce diverse data and information, which can be learned using ML to extract and purify knowledge for further enhancing their performance. It can be predicted that learnable MOEAs will play an essential role in handling complex MOPs in the future development of MOEAs. A closer inspection of the most reputed scientific databases also unveils those efforts carried out in learnable MOEAs are exponentially growing in recent years. This upsurge of studies demands insightful reference material to 1) summarize related achievements so far, 2) detect and analyze research trends, 3) discuss current limitations and perform a profound reflection, and 4) prescribe some interesting future research directions. This is the rationale for this survey. Thus, a comprehensive review of existing efforts on learnable MOEAs for scaling-up MOPs is proposed from the following five aspects:
\begin{enumerate}
\item{General definition and taxonomy of scaling-up MOPs and learnable MOEAs;}
\item{Efforts on learnable evolutionary discriminators for handling the MOPs with a scaling-up objective space, i.e., many-objective optimization problems (MaOPs);}
\item{Efforts on learnable evolutionary generators for solving the MOPs with a scaling-up search space, i.e., large-scale MOPs (LMOPs);}
\item{Efforts on learnable evaluators to alleviate the challenges posed by MOPs with the scaling-up cost of function evaluations, i.e., expensive MOPs (EMOPs);}
\item{Efforts on learnable evolutionary transfer modules for finding a shortcut that can improve the efficacy when optimizing a scaling-up number of different MOPs sequentially or simultaneously, i.e., sequential MOPs (SMOPs) and multitasking MOPs (MMOPs).}
\end{enumerate}
It is worth pointing out that many related submissions have been made available publicly, but there is no comprehensive survey of the literature on learnable MOEAs. Although a review on evolutionary computation plus ML has been made in \cite{zhang2011evolutionary}, it mainly focuses on reviewing different ML methods applied to assist swarm intelligence algorithms, rather than concentrating on the learnable MOEAs. Besides, most of the references in \cite{zhang2011evolutionary} are pre-2012 and do not include an update on the papers published during the past ten years for learnable MOEAs. This paper presents a survey involving a myriad of new learnable MOEA studies, with the expectation to inspire some new ideas for promoting the development of MOEA in solving complex MOPs.

The rest of this paper is organized as follows. We first provide the general definition and taxonomy of scalable and learnable evolutionary multiobjective optimization (EMO) in Section II. Then, we introduce the existing efforts with respect to learnable discriminators, learnable generators, learnable evaluators, and learnable transfer modules in Sections III-VI. Finally, we conclude this paper and provide some potential research directions in Sections VII-VIII.
\begin{figure*}[!t]
\centering
\includegraphics[width=6.2in]{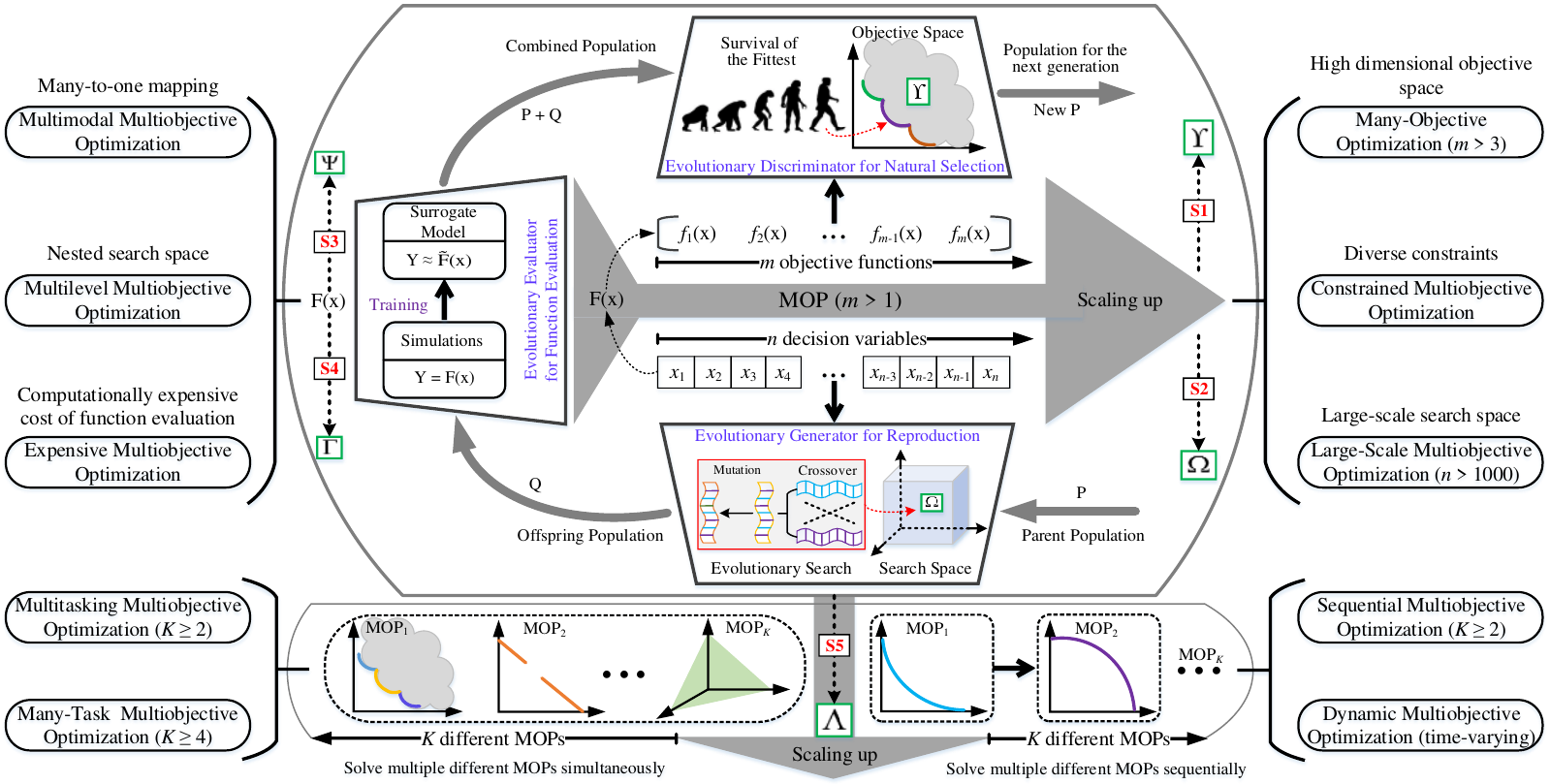}
\caption{Taxonomy of scaling-up MOPs according to the struct $S_{\mathrm{MOP}}$ from its five scalable components, including S1: objective space $\Upsilon$, S2: search space $\Omega$, S3: mapping diagram $\Psi$, S4: computational cost $\Gamma$ for function evaluation, and S5: problem domain $\Lambda$.}
\label{fig_sim}
\end{figure*}
\section{Scalable and Learnable EMO}
This section briefly introduces the preliminaries related to scalable EMO, including the general definition and taxonomy of scaling-up MOPs and the challenges they bring. Accordingly, we present the taxonomy of learnable MOEAs.
\subsection{Scalable Multiobjective Optimization Problems}
An MOP comprises multiple objective functions to be minimized simultaneously, which can be formulated as follows:
\begin{equation}
\begin{aligned}
\operatorname{Minimize} & \mathrm{F}(\mathrm{x})=\left(f_1(\mathrm{x}), \ldots, f_m(\mathrm{x})\right), \\
& \text { s.t. } \mathrm{x} \in \Omega, \mathrm{F}(\mathrm{x}) \in \Upsilon
\end{aligned}
\end{equation}
where $\mathrm{x} =\left(x_1, \ldots, x_n\right)$ is a solution vector with $n$ often inter-related variables from the search space $\Omega$, and $\mathrm{F}(\mathrm{x})$ defines $m$ often mutually conflicted functions $f_1(\mathrm{x}), \ldots, f_m(\mathrm{x})$ in the objective space $\Upsilon$, none of which can be preferred over others $(m \geq 2)$. Therefore, without additional subjective preference, we cannot find a single optimal solution to simultaneously optimize all objectives of an MOP, but a (possibly infinite) set of Pareto optimal solutions, called the Pareto set (PS). The mapping of PS in the objective space is called PF. Thus, the goal of solving an MOP by MOEAs may be to find a representative set of solutions that can evenly and closely approximate the PS/PF \cite{zhan2022survey}, or to find a single solution that satisfies the subjective preferences of a decision-maker \cite{duro2014machine}. In this paper, a struct $S_{\mathrm{MOP}}$, including five scalable components, is defined to express the scalability of MOPs, as follows:
\begin{equation}
S_{\mathrm{MOP}}=(\Omega, \Upsilon, \Psi, \Gamma, \Lambda)
\end{equation}
where $\Upsilon$ and $\Omega$ respectively represent the objective space and the search space of an MOP $\in \Lambda$; $\Psi$ is the mapping diagram of $\mathrm{x} \rightarrow \mathrm{F}(\mathrm{x})$ in (1) for this MOP; $\Gamma$ denotes the computational cost of evaluating $\mathrm{F}(\mathrm{x})$; and $\Lambda$ indicates the problem domain including $K$ different MOPs to be solved, $K \geq 1$. According to this struct $S_{\mathrm{MOP}}$, the taxonomy of scaling-up MOPs is illustrated in Fig. 2. The details are elaborated below.

From the aspect of scalable $\Upsilon$, an MOP with $m$ = 2 or 3 can be scaled up to become an MaOP with $m \geq 4$. Since constraints can be reformulated as optimization objectives, the MOP with scaling-up constraints can be implicitly regarded as an MaOP. Researchers interested in constrained EMO can refer to its latest survey \cite{liang2022survey}. Considering the scalable $\Omega$, an MOP with a small-scale $\Omega$ can be switched to an LMOP with a large-scale $\Omega$, in which its dimensionality $n$ scales up to more than one thousand, i.e., $n \geq 1000$ \cite{tian2021evolutionary}.

Regarding the scalable $\Psi$, an MOP with a one-to-one mapping can be scaled up to become a multimodal MOP with a many-to-one mapping, in which multiple diverse optimal variable vectors in $\Omega$ may share the nearly same objective vector in $\Upsilon$. Besides, most MOPs follow direct mapping, i.e., the vector $\mathrm{x}$ is directly paired to the function vector $\mathrm{F}(\mathrm{x})$ without involving any nested mapping. However, such direct mapping does not reflect the complexity level of practical problems, where the mapping may be modulated through multiple levels of indirection, e.g., bi-level MOP with a nested search space. Interested readers can refer to the latest surveys \cite{tanabe2020review} and \cite{sinha2017review} respectively for multimodal EMO and bi-level EMO. Since the $\mathrm{F}(\mathrm{x})$ may be unexpressed and noisy in practice, its evaluation will become more and more computationally expensive as the fidelity requirement increases \cite{jin2019data}. Thus, from the aspect of implicitly scalable $\Gamma$, an inexpensive MOP can be scaled up to become an expensive MOP (EMOP), which requires prohibitive computational cost for function evaluations.

The growing complexity of MOPs (e.g., MaOPs, LMOPs, and EMOPs) and real-world MOPs hardly appear in isolation have revealed the necessity of sharing optimization experience by processing a scaling-up number of $K$ different MOPs in sequentially or simultaneously \cite{ong2016evolutionary}. Accordingly, the resultant scaling-up problem is an SMOP or an MMOP with $K \geq 2$. Particularly, if we treat an MOP as a static state, then an SMOP will have multiple different states. When this sequential switch from one state to the new one can be simulated as a time-varying process, the resultant problem is a dynamic MOP (DMOP) \cite{yazdani2021survey}. Thus, a DMOP can be implicitly regarded as a sequential many-task MOP \cite{shakeri2022scalable}, in which all states are solved sequentially and the MOP in each state can be solved with the help of the knowledge or experiences accumulatively learned from previous optimization exercises \cite{liu2022DRN}.
% \usepackage[normalem]{ulem}
\begin{table*}[]
\caption{Summarization of the Bottleneck for Traditional MOEAs in Solving Scaling-up MOPs\label{tab:table1}}
\centering
\begin{tabular}{l|l|l|l|l}
\hline
\multicolumn{1}{c|}{Aspects}                                                                                    & \multicolumn{1}{c|}{Scale-up MOPs}                                                                                                      & \multicolumn{1}{c|}{Main modules}                                                                                                                                               & \multicolumn{1}{c|}{Basic strategies}                                                                                                                                                                                                      & \multicolumn{1}{c}{Challenges or limitations}                                                                                                                                                                                                                                                                                                                                                                                                                                                           \\ \hline
\begin{tabular}[c]{@{}l@{}}$\mathrm{S}1-\Upsilon$:\\ objective\\ space\end{tabular}                             & \begin{tabular}[c]{@{}l@{}}$\diamond$ MaOPs\\ (many-objective)\end{tabular}                                                             & \begin{tabular}[c]{@{}l@{}}$\checkmark$ Discriminator;\\ //outputs survival\\ solutions by filtering\\ poorly performed\\ solutions from the\\ inputted population\end{tabular} & \begin{tabular}[c]{@{}l@{}}Selection strategies:\\ $\oplus$ Pareto-based \cite{deb2002fast};\\ $\oplus$ indicator-based \cite{shang2020survey};\\ $\oplus$ decomposition-based \cite{li2021decomposition};\\ $\oplus$ their hybridization \cite{sindhya2012hybrid};\end{tabular}                     & \begin{tabular}[c]{@{}l@{}}$\odot$ insufficient selection pressure for Pareto-based\\ strategies (mutually non-dominated solutions) \cite{deb2013evolutionary}; \\ $\odot$ efficacy of decomposition-based strategies\\ deteriorates severely as pre-specified weights \\ cannot match irregular PFs well \cite{ishibuchi2016performance}; \\ $\odot$ cost of computing performance indicators\\ grows exponentially as objective increases \cite{deng2019approximating}.\end{tabular}                                                                               \\ \hline
\begin{tabular}[c]{@{}l@{}}$\mathrm{S}2-\Omega$:\\ search\\ space\end{tabular}                                  & \begin{tabular}[c]{@{}l@{}}$\diamond$ LMOPs\\ (large-scale)\end{tabular}                                                                & \begin{tabular}[c]{@{}l@{}}$\checkmark$ Generator; \\ //outputs offspring\\ by finding promising\\ solutions in search\\ space based on the\\ inputted parents\end{tabular}     & \begin{tabular}[c]{@{}l@{}}Search strategies:\\ $\oplus$ mutation;\\ $\oplus$ crossover;\\ $\oplus$ particle swarm optimizer;\\ $\oplus$ differential evolution;\\ $\oplus$ evolution strategy \cite{zhang2020evolution};\\ $\oplus$ etc.\end{tabular}      & \begin{tabular}[c]{@{}l@{}}$\odot$ search space expands exponentially, requiring\\ generator with growing search capability \cite{liu2021LMF};\\ $\odot$ interaction between variables are more intricate;\\ $\odot$ contribution of variables is more imbalanced;\\ $\odot$ overlap between shared variables that define\\ different objectives becomes more disorganized \cite{zapotecas2018review};\\ $\odot$ overly-increasing number of variables will\\ inevitably make function evaluation expensive \cite{wang2017generic}.\end{tabular} \\ \hline
\begin{tabular}[c]{@{}l@{}}$\mathrm{S}3-\Psi$:\\ mapping\\ diagram\end{tabular}                                 & \begin{tabular}[c]{@{}l@{}}$\diamond$ MMMOPs\\ (multimodal)\\ $\diamond$ BMOPs\\ (bi-level)\end{tabular}                                & \begin{tabular}[c]{@{}l@{}}$\checkmark$ Discriminator;\\ $\checkmark$ Generator;\end{tabular}                                                                                   & \begin{tabular}[c]{@{}l@{}}Integrated strategies:\\ $\oplus$ diversity strategies\\ cover objective \& search space;\\ $\oplus$ search strategies cover\\ low- \& up-level variables.\end{tabular}                                         & \begin{tabular}[c]{@{}l@{}}$\odot$ difficult to maintain diversity in both objective \&\\ search space of multimodal cases \cite{tanabe2020review};\\ $\odot$ difficult to find Pareto solutions with both optimal\\ upper- \& lower-level variables \cite{sinha2017review}.\end{tabular}                                                                                                                                                                                                                                            \\ \hline
\begin{tabular}[c]{@{}l@{}}$\mathrm{S}4-\Gamma$:\\ computation\\ cost for \\ function\\ evaluation\end{tabular} & \begin{tabular}[c]{@{}l@{}}$\diamond$ EMOPs\\ (expensive)\end{tabular}                                                                  & \begin{tabular}[c]{@{}l@{}}$\checkmark$ Evaluator;\\ //outputs objective\\ function values or\\ fitness values\\ for each inputted\\ solution\end{tabular}                      & \begin{tabular}[c]{@{}l@{}}Evaluation strategies:\\ $\oplus$ evaluate objectives\\ from given expressions;\\ $\oplus$ predict objectives by\\ surrogates learned from\\ simulation data when they\\ are unexpressed or noisy.\end{tabular} & \begin{tabular}[c]{@{}l@{}}$\odot$ require prohibitive computing cost by evaluating\\ objective functions directly \cite{durillo2010study};\\ $\odot$ construct proper surrogate models require many\\ promising solutions \cite{jin2008pareto};\\ $\odot$ simulation data is often imbalanced \& rough in\\ unexpressed or noisy cases \cite{jin2019data}; \\ $\odot$ approximation errors are increased accordingly,\\ which may mislead search \& selection \cite{jin2019data}.\end{tabular}                                                        \\ \hline
\begin{tabular}[c]{@{}l@{}}$\mathrm{S}5-\Lambda$:\\ problem\\ domain\end{tabular}                               & \begin{tabular}[c]{@{}l@{}}$\diamond$ MMOPs\\ (multitask)\\ $\diamond$ SMOPs\\ (sequential)\\ $\diamond$ DMOPs\\ (dynamic)\end{tabular} & \begin{tabular}[c]{@{}l@{}}$\checkmark$ Discriminator;\\ $\checkmark$ Generator;\\ $\checkmark$ Evaluator.\end{tabular}                                                         & \begin{tabular}[c]{@{}l@{}}Transfer strategies:\\ $\oplus$ without transfer by\\ solving MOPs independently;\\ $\oplus$ implicit transfer by\\ assortative mating and \\ selective imitation \cite{gupta2016multiobjective, bali2020cognizant, zhou2020toward};\end{tabular}               & \begin{tabular}[c]{@{}l@{}}$\odot$ start from scratch when facing new coming MOPs,\\ regardless of related optimization experience;\\ $\odot$ generalization abilities of traditional MOEAs are \\ so poor that cannot leverage experience to achieve\\ more promising optimization efficacy;\\ $\odot$ implicit knowledge transfer with random \& blind\\ transfer fails to improve overall performance.\end{tabular}                                                                                      \\ \hline
\end{tabular}
\end{table*}
\subsection{The Bottleneck of MOEAs for Scaling-up MOPs}
Given an MOP to be solved, the appointed MOEA will iteratively run a stochastic optimization process on a population $P$ to approximate the optima. As shown in Fig. 2, this process mainly includes three modules: 1) the generator for reproduction that reproduces a new offspring population $Q$ by exploring the search space; 2) the function evaluator for calculating the objective values for each solution; 3) the discriminator for environmental selection that keeps the fittest solutions from the combined population $P+Q$ to the next generation. The potential challenges that circumscribe traditional MOEAs' generalization ability in handling scaling-up MOPs are summarized in Table I.

Traditional MOEAs face great challenges in responding to the ever-growing dimensionality and complexity of MOPs. These scaling-up MOPs spring up exuberantly in practical applications, e.g., in deep learning models with ever-expanding architecture and engineering systems with ever-complex linkage \cite{zhai2014emerging, zhou2014big}. Thus, to continue being applicable to realist, MOEAs should be equipped with learning ability to cope with new challenges brought by these scaling-up MOPs. The problem-specific features, if well learned with ML techniques, can be a great help to enhance the adaptive ability \cite{li2021survey} and generalization ability \cite{bandaru2017data} of MOEAs.

\subsection{Learnable Multiobjective Evolutionary Algorithms}
In classic MOEAs, their generators and discriminators are developed based on some traditional genetic operators (e.g., crossover, mutation, and selection). These two evolutionary modules are unable to learn the exact characteristics of an MOP they encounter. Thus, they do not have the ability to respond flexibly to the potential challenges of solving this black-box MOP. Besides, most MOEAs are designed on the basis that all objective function expressions are known. However, an MOP may be unexpressed and noisy in practice. Without using additional measures to predict the exact expression in these uncertain cases, MOEAs are not able to work satisfactorily \cite{jin2019data}. In addition, traditional MOEAs solve all MOPs independently, which is not an efficient manner when solving multiple MOPs sequentially or simultaneously. Many existing studies have shown that optimization efficiency can be significantly improved by transferring knowledge acquired from solving different (but somehow related) MOPs.
\begin{figure*}[!t]
\centering
\includegraphics[width=6.2in]{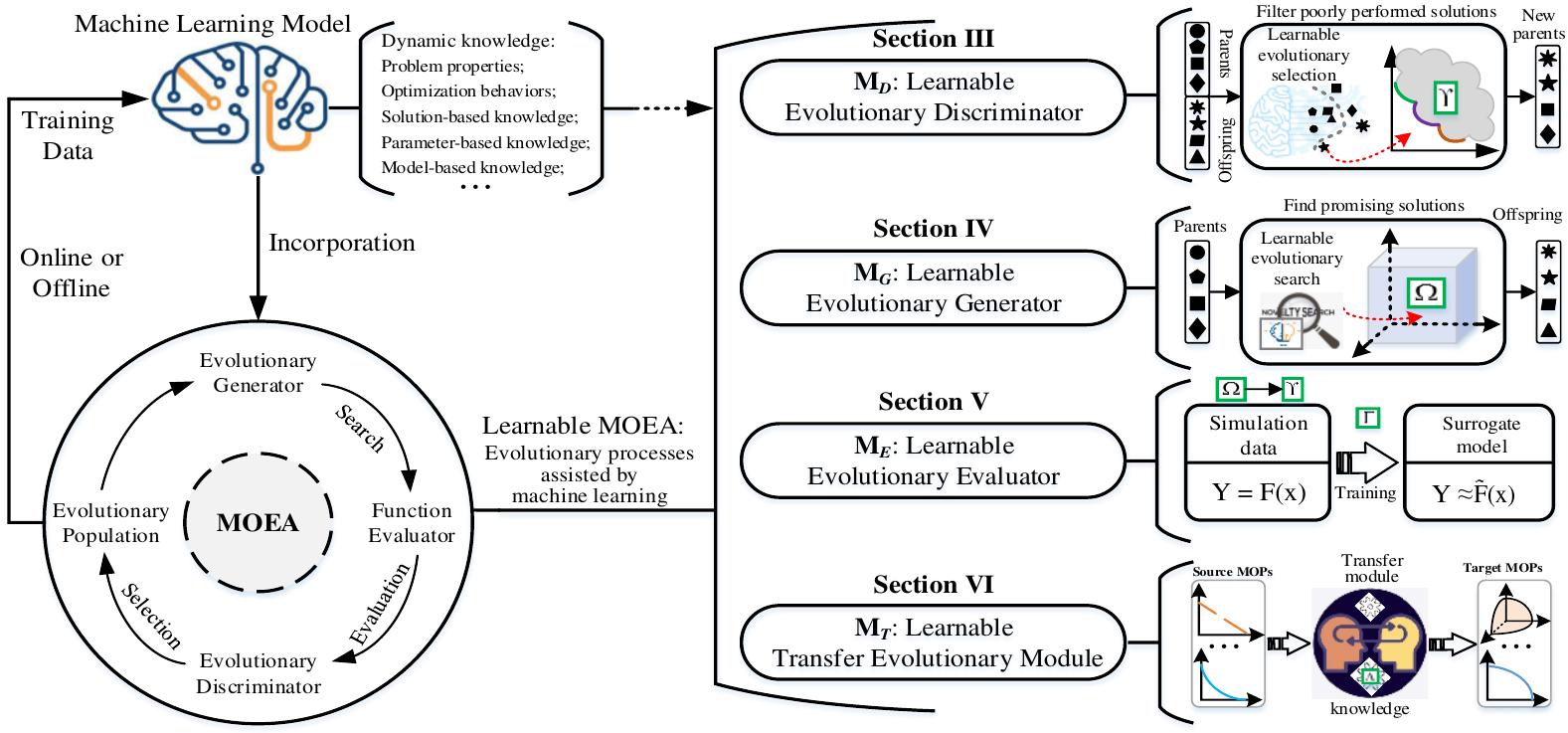}
\caption{Taxonomy of learnable MOEAs according to $L_{\mathrm{MOEA}}$ from its four learnable components, including $\mathrm{M}_D$: evolutionary discriminator, $\mathrm{M}_G$: evolutionary generator, $\mathrm{M}_E$: evolutionary evaluator, and $\mathrm{M}_T$: evolutionary transfer modules.}
\label{fig_sim}
\end{figure*}
In this context, learnable MOEAs are coupled with various ML techniques to assist their evolutionary processes so that they are capable of learning \cite{michalski2000learnable, cheng2018model}. Generally, learnable MOEAs build ML models to acquire potentially useful knowledge, e.g., problem properties, optimization behaviors, solution-based knowledge, parameter-based knowledge, model-based knowledge \cite{liu2022DRN}, etc. With proper guidance from the learned knowledge, they aim to improve the selection performance of its discriminator and the search capability of its generator \cite{lin2018hybrid}. When facing expensive MOPs, they expect to lightly predict the objective values of these MOPs using surrogate models, which are trained on the available simulation data collected from $\Omega$ to $\Upsilon$. Moreover, it is sensible to create a transfer learning module between different MOP domains (often divided into source domains and target domains). The knowledge learned from solving source MOPs can be transferred to assist the optimization of target MOPs in a more efficient manner \cite{tan2021evolutionary}.

To clearly describe the ``learning capability'' of MOEAs, a struct $L_{\mathrm{MOEA}}$ with four learnable modules is defined below:
\begin{equation}
L_{\mathrm{MOEA}}=\left(\mathrm{M}_D, \mathrm{M}_G, \mathrm{M}_E, \mathrm{M}_T\right)
\end{equation}
where $\mathrm{M}_D$, $\mathrm{M}_G$, and $\mathrm{M}_E$ are respectively the discriminator, generator, and evaluator of this MOEA; $\mathrm{M}_T$ indicates a learnable transfer module when this MOEA is used to solve scaling-up MOPs from the aspect of $\Lambda$. Thus, according to $L_{\mathrm{MOEA}}$, existing learnable MOEAs can be roughly classified into four categories: MOEA with a learnable discriminator, MOEA with a learnable generator, MOEA with a learnable evaluator, and MOEA with a learnable transfer module, as shown in Fig. 3. Their imperatives and the main scaling-up MOPs they are expected to solve are briefly described below.

\textbf{Learnable evolutionary discriminators:} ML techniques are used to subtly assist the environmental selection, which can handle the potential challenges brought by MaOPs with scaling-up objective spaces. With the increasing number of objectives, the ability to discriminate poor and elite solutions of existing environmental selection strategies deteriorates seriously. Specifically, the discriminator's convergence pressure is insufficient and its diversity maintenance becomes unsatisfactory. Thus, learning how to improve the discriminative ability of selection strategies becomes particularly essential for solving MaOPs. The efforts made in this regard will be elaborately reviewed in Section III.

\textbf{Learnable evolutionary generators:} ML techniques are used to tactfully assist the evolutionary search, which can adaptively respond to the potential challenges brought by LMOPs with large-scale search spaces. The search capability of most existing genetic operators deteriorates dramatically in the large-scale search space, which leads to the fact that the ineffective offspring are often reproduced by the generator. Thus, learning how to enhance the search capability of the generators becomes extremely important for solving LMOPs. The efforts made in this regard will be comprehensively surveyed in Section IV.

\textbf{Learnable evolutionary evaluators:} Discriminators are customized based on solutions' objective values, which are outputted by an evaluator. If the evaluator is blocked, e.g., in solving unexpressed MOPs, the discriminator will be unbootable. It is worth noting that the evaluator can be regarded as a discriminator to directly filter poorly performed solutions for single-objective optimization problems. In other words, the discriminator and evaluator are the same module in single-objective EAs, called fitness evaluator. To highlight the differences, this paper discusses them separately. The details of learnable evolutionary evaluators are provided in Section V.

\textbf{Learnable transfer evolutionary modules:} evolutionary modules via transfer learning are proposed to share useful optimization knowledge between source-target MOPs. Intuitively, it is a shortcut to solving the target MOP via reusing the experiences learned from the optimization exercises of solving source problems. The optimization of source MOPs can be completed or carried out at the same time as the target MOPs, and the respective resultant forms of transfer optimization are sequential form and multitasking form \cite{liu2022DRN}. The target MOPs at hand are sequentially solved with the help of the experiences accumulatively learned from previous optimization exercises in the sequential form. By contrast, all the MOPs, starting from scratch, are optimized simultaneously in the multitasking form. The efforts made in learnable transfer evolutionary modules will be systematically investigated in Section VI.

\section{Learnable Evolutionary Discriminators}
In an MOEA, its discriminator is customized to select the next population with balanceable convergence and diversity from the combined population of parent and offspring. As shown in Fig. 4, this section starts by answering what kind of dynamic knowledge the discriminator wants to learn in the objective space. Then, a review of existing efforts on learnable discriminators is given according to the types of ML methods used to acquire knowledge. Here, three traditional types of ML (i.e., supervised, unsupervised, and reinforcement learning) methods are mainly considered by the discriminator to improve its performance of environmental selection.

\subsection{Learnable Discriminators via Supervised Learning}
Supervised learning is defined as using labeled datasets to train a model or function that classifies future data or predicts outcomes accurately. With training data fed into the model, its parameters are dynamically adjusted until the model has been fitted appropriately. The dataset, used to train a supervised learning model, consists of the objective vectors of all candidate solutions in $\Upsilon$. Intuitively, a classifier model can be directly learned to predict the Pareto-rank of solutions in Pareto-based selections \cite{chang2021self}, to approximate the hypervolume contributions of solutions in indicator-based selections \cite{deist2021multi}, and to evaluate the meta quality of solutions in decomposition-based selections \cite{chen2017decomposition}. However, these selection strategies of directly using a classifier model to filter solutions are often challenging when solving MaOPs with scaling-up objective spaces \cite{zhang2022classification}. In this context, the relatively attractive way to assist the discriminator via supervised learning for handling MaOPs, especially for those with irregular PFs, is using an $m-1$ dimensional simplex model, which can dynamically approximate the geometry shape of their PFs.

In general, the geometry of the non-dominated front is estimated by training a regression model. Specifically, in some MOEAs \cite{liu2010t,zapotecas2014using}, the geometry of the non-dominated front can be predefined as the following simple curve:
\begin{equation}
\left(f_1^{\prime}(\mathrm{x})\right)^p+\ldots+\left(f_i^{\prime}(\mathrm{x})\right)^p+\ldots+\left(f_m^{\prime}(\mathrm{x})\right)^p=1, p>0
\end{equation}
where $f_1^{\prime}(\mathrm{x})$ indicates the $i$-th normalized objective value of a solution $\mathrm{x}$, which is generally calculated as follows:
\begin{equation}
f_i^{\prime}(\mathrm{x})=\frac{f_i^{\prime}(\mathrm{x})-z_i^{\min }}{z_i^{\max }-z_i^{\min }}
\end{equation}
where $z^{\min }$ and $z^{\max }$ are respectively the ideal and nadir points of the population in historical memory, $i=1,2, \ldots, m$, and $m$ represents the number of objectives. Obviously, this family of surfaces in (4) involves only one parameter (i.e., $p$) that needs to be learned, which reflects the curvature of the PF. Given a non-dominated solution set, termed $NS$, the value of $p$ can be directly estimated by the least square fitting method as follows:
\begin{equation}
p=\underset{p}{\arg \min } \sum_{\mathrm{x} \in N S}\left[\left(f_1^{\prime}(\mathrm{x})\right)^p+\ldots+\left(f_m^{\prime}(\mathrm{x})\right)^p-1\right]^2
\end{equation}
To reduce the computational cost, the prediction of $p$ is limited to a predetermined and finite range of discrete values in some following works \cite{liu2020MaOEAAC,luo2018approximating}. Moreover, in \cite{zhou2009approximating}, the PF of an MOP is assumed as the following linear model:
\begin{equation}
a_1\left(f_1^{\prime}(\mathrm{x})\right)^{p_1}+\ldots+a_i\left(f_i^{\prime}(\mathrm{x})\right)^{p_i}+\ldots+a_m\left(f_m^{\prime}(\mathrm{x})\right)^{p_m}=1
\end{equation}
where the hyperparameters are learned by maximizing the log marginal likelihood of Gaussian process regression in \cite{wu2018learning} and by the Levenberg-Marquardt algorithm in \cite{tian2018guiding}.

In this way, the learned PF's geometric properties (e.g., curvature, concavity, and convexity) can be used to guide the environmental selection for solving irregular MaOPs. Specifically, weights can be sampled adaptively \cite{siwei2011multiobjective} and scalarizing methods can be tailored properly according to the learned curvature of the PFs \cite{wang2016decomposition}. Thus, both the selection pressure and the maintenance of population diversity can be improved in decomposition-based strategies, as studied in learnable decomposition \cite{wu2018learning} and fuzzy decomposition \cite{liu2020fuzzy} of MaOPs. According to the predicted model of the PF, adaptive performance indicators can be defined to properly evaluate the quality (convergence or diversity) of solutions in indicator-based selections, e.g., generic front model-based performance criteria \cite{tian2018guiding}, Minkowski distance \cite{xu2018evolutionary}, and direction similarity between solutions \cite{liu2020fuzzy}. Furthermore, a Pareto-adaptive dominance (e.g., $pa \epsilon$-dominance \cite{hernandez2007pareto}) can be defined and Pareto regions can be learned \cite{rebello2021pareto} to improve the selection pressure of Pareto-based strategies with the help of predicted geometries.

These simplex model-assisted selection strategies have been verified to be effective in solving MaOPs with irregular continuous PFs \cite{tian2022local}. However, it is still a challenging task to maintain population diversity on MaOPs with disconnected, degenerated, and inverted PFs. To address this issue, some unsupervised learning-assisted selection strategies are proposed to enhance the performance consistency of MOEAs for tackling MaOPs with various PF shapes. The details are reviewed as follows.

\subsection{Learnable Discriminators via Unsupervised Learning}
Unsupervised learning uses ML algorithms to analyze and cluster unlabeled datasets. These algorithms discover hidden patterns or data distributions without the need of human intervention. Intuitively, clustering methods can be used to divide the population into multiple clusters in the objective space. The distribution information of solutions can be preserved through these clusters. The clustering results can assist the discriminators to select a population with balanceable convergence and diversity when handling MaOPs \cite{hua2021survey}.
\begin{figure*}[!t]
\centering
\includegraphics[width=6.4in]{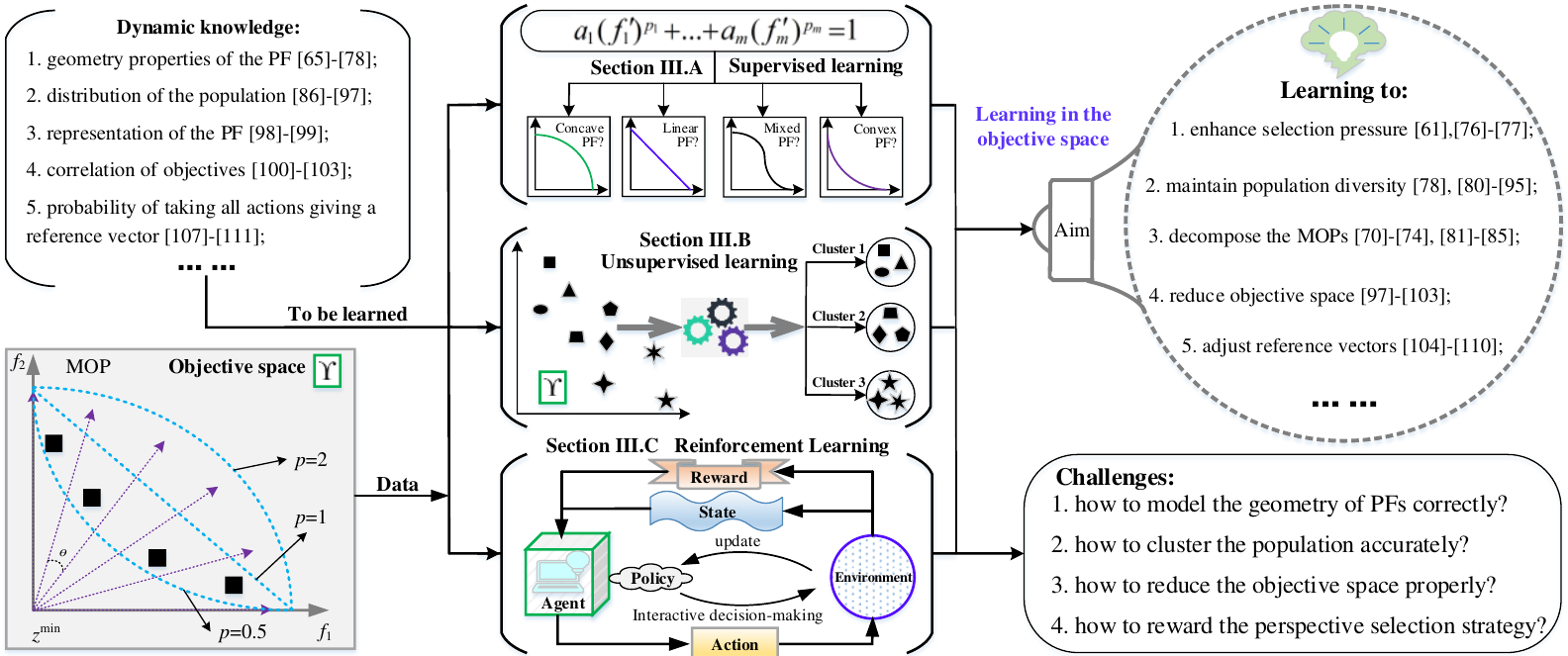}
\caption{Illustrating what knowledge can be learned in the objective space via various ML models and the aims of learning this dynamic knowledge.}
\label{fig_sim}
\end{figure*}
In the early Pareto-based strategies \cite{zitzler1999multiobjective}, the clustering methods with a Euclidean metric are used to prune the population when the size of the nondominated set exceeds preset bounds, which can make up the population diversity. However, these clustering-based pruning operators show insufficient selection pressure on scaling-up MaOPs as almost all solutions are mutually non-dominated from the beginning of evolution. In this context, decomposition-based strategies seem to stand out in terms of selection pressures for solving MaOPs. Various decomposition-based strategies are developed to divide the combined population into multiple mutually exclusive clusters according to a set of predefined reference vectors. Here, each solution belongs only to the cluster specified by its nearest reference vector \cite{liu2020self}. Then, some representative solutions with better convergence are selected from each cluster. To be specific, only the solutions belonging to the same cluster are compared based on the adopted convergence indicators, e.g., locally weighted sum \cite{wang2016localized}, $\theta$-dominance \cite{yuan2015new}, angle-penalized distance \cite{cheng2016reference}, and clustering-ranking \cite{cai2015clustering}. This constrained decomposition can be analogous to partition-based clustering with a direction similarity metric (e.g., the distance between two direction vectors), in which reference vectors are regarded as fixed centroids \cite{liu2020self}.

The main limitation of these fixed clustering-assisted selections is that their performance is strongly dependent on the matching between preset reference vectors and the target PF \cite{ishibuchi2016performance}. To circumvent this limitation, some up-and-coming strategies divide the population into multiple clusters via iteratively dynamic clustering methods without using reference vectors. For example, hierarchical clustering is used in \cite{lin2018clustering,sun2018learning,liu2019novel,hua2018clustering,denysiuk2014clustering}, $k$-means and fuzzy $c$-means are adopted in \cite{bejarano2022clustering}, the Gaussian mixture model is applied in \cite{li2019multiobjective}, density-based clustering is employed in \cite{kramer2010dbscan}, and cascade clustering is used in \cite{ge2018many}. These dynamically obtained clusters can match the geometric properties of the target PF adaptively. Concretely, the population is dynamically partitioned into multiple non-empty clusters for preserving population diversity, where solutions with higher similarity are gathered into the same cluster. Then, the representative solution of each cluster is survived into the next generation to maintain population convergence.

Nevertheless, dynamic clustering methods need to iteratively run many times to obtain final clusters with more balanceable diversity and convergence \cite{liu2021evolutionary}. So there are three key concerns to be noted in designing dynamic clustering selections: 1) how to select a suitable clustering method? Although hierarchical clustering is commonly used to assist environmental selection, few efforts explain why it is the most appropriate. 2) How to select the similarity metric? Using Euclidean distance for clustering may lead to dominance-resistant solutions \cite{liu2017many} being divided into a separate cluster. They have a high probability of surviving to the next generation, which will prevent the population from moving towards PF. To alleviate this issue, cosine similarity is widely used for clustering the population. The fairness of cosine similarity defined based on the ideal point is still dependent on the curvature of the target PF. It is often necessary to predict the curvature of PF in advance and then define the corresponding cosine similarity \cite{liu2021evolutionary}. 3) How to select representative solutions in each cluster? The solution with the best convergence is always chosen as the representative of each cluster. Thus, it becomes critical to properly evaluate the convergence of solutions clustered in a locally constrained space.

In addition, reference vectors can be extracted adaptively by clustering to improve the performance of decomposition-based selection strategies \cite{liu2020self}. The dimensionality of objective space can be reduced by clustering according to the conflict between objectives \cite{pal2018decor}. Intuitively, learning to reduce the number of objectives would yield shorter running times of selection, especially for hypervolume-based selection \cite{brockhoff2007improving}. Meanwhile, an MaOP can be transferred to an MOP by eliminating objectives that are not essential to describe the PF, followed by solving it with a typical MOEA \cite{yuan2017objective}. Inspired by this, various dimensionality reduction methods, like principal component analysis (PCA) \cite{deb2006searching}, semidefinite embedding \cite{saxena2012objective}, and transfer matrix with kriging model \cite{ma2020novel}, are applied to reduce the objective space of MaOPs. Although the reduced representation of an MaOP will have a favorable impact on the selection efficiency, there are still three core concerns worth discussing and analyzing for objective space reduction \cite{brockhoff2009objective}:
\begin{enumerate}
\item{Whether it is possible to ignore some objectives while describing the PF of the target MaOP? and under what conditions are such objective reductions feasible?}
\item{How to define the correlation and conflict among objectives reasonably and quantitatively? and find out the insignificant or redundant objectives accordingly;}
\item{How to compute a minimum set of objectives ac-cording to such qualitative objective space changes?}
\end{enumerate}
Furthermore, adaptive reference adjustment via other unsupervised learning methods, e.g., self-organizing map \cite{gu2017self} and growing neural gas \cite{liu2020adaptive,liu2019adapting}, is also a considerable way to promote decomposition-based selections in solving MaOPs.

\subsection{Learnable Discriminators via Reinforcement Learning}
Reinforcement learning (RL) aims to learn the probability of taking an action, which will maximize the expected cumulative reward at the current state. Concretely, the agent takes an action at the current state based on the output of the policy network. Then, a reward is obtained after sending the action to the environment, followed by switching to the next cycle of inputting the new state and reward into the agent. This process will repeat until a terminal condition is fulfilled, as shown in Fig. 4. To solve a combinational MOP, the RL technique is applied in \cite{li2020deep}. To be specific, an MOP is first divided into multiple subproblems with a set of reference vectors in the objective space, and then each sub-problem is modeled as a neural network. Subsequently, the model parameters of all neural networks are optimized by a deep RL algorithm. The optimal solutions can be directly obtained via these optimized models.

The key to improving the performance of decomposition-based discriminators is to redress the mismatch between the reference vectors and the PF shape. Given a finite number of actions, the most appropriate action can be selected to adjust the reference vector according to its current state and the probability of performing each action. This process can be realized by RL techniques. Inspired by this, an adaptive reference vector approach via RL is proposed in \cite{ma2021learning}, in which the action space is defined as [\textit{regeneration}, \textit{relocation}, \textit{return-back}, \textit{state-keeping}] for the agent to control its reference vector. Besides, each agent of the reference vector can receive five different states according to the quantity of its associated dominated and non-dominated solutions. Finally, the Q-learning method \cite{cheng2022multi} is used to learn the optimal policy network. An RL-assisted decomposition-based discriminator is also developed in \cite{zhang2021modrl}. To the best of our knowledge, there are few efforts to improve the performance of environmental selection via RL, but this does not mean it is not a promising research direction. For example, RL techniques have been used to automatically design an appropriate discriminator and set optimal parameters when facing different MaOPs \cite{zhang2022meta}.

\section{Learnable Evolutionary Generators}
With the inputted parent solutions, an evolutionary generator aims to reproduce offspring solutions that can move in the promising direction to approach the unknown PS fast. To achieve this, it explores the search space by processing parent solutions via genetic operators. However, the randomness-based genetic operators are often unable to grasp the direction and pace of the search, which may drive the population to move forward roughly and slowly. Consequently, greatly high amounts of computing resources are required to approach the PS when solving LMOPs. As computing budgets are often limited in practice, it becomes particularly essential for a generator to simplify the large-scale search space or guide the population to converge fast along with the promising directions (e.g., improve its search capability). In this context, ML techniques are widely used to discover the structural information of search space and the latent promising converging paths of solutions excavated from available data \cite{deb2012hybrid}. The learned model can then be applied to find ways of guiding evolutionary search with high efficiency in responding to the challenges brought by the scaling-up LMOPs.
%See \cite{ref1,ref2,ref3,ref4,ref5} for resources on formatting math into text and additional help in working with \LaTeX .

This section will start with a brief review of learnable generators in multiobjective estimation of distribution algorithms (MEDAs), which directly learn a probabilistic model to sample solutions. The limitations of MEDAs in solving scaling-up LMOPs are discussed thereafter. Then, the following survey route is outlined in Fig. 5. Similarly, this section focuses on three main types of ML (i.e., supervised, unsupervised, and RL) methods and investigates how they can be used to enhance the search capability of genetic operators.

\subsection{Learnable Generators via Estimation of Distribution}
To mitigate the adverse influence (e.g., the disruption of building blocks and mating restrictions) led by genetic operators, the evolutionary generator is directly designed based on a probabilistic model built to approximate the PS in most MEDAs \cite{hauschild2011introduction}. Here, the probabilistic model can be a Gaussian distribution \cite{dong2013scaling}, a Bayesian network \cite{laumanns2002bayesian}, a regression decision tree \cite{zhong2007decision}, a restricted Boltzmann machine \cite{tang2010restricted}, a growing neural gas \cite{marti2011mb}, a manifold learning model \cite{li2014general}, etc. In this way, new solutions can be directly sampled with the probabilistic model. It is progressively learned from the statistic information (e.g., the dependencies between variables and the regularity property of the PS), which is captured from the existing solutions. Clearly, the model-building is the key step of MEDAs, which leads to the following four main classes of MEDAs based on the adopted modeling strategies:
\begin{enumerate}
\item{Graphical model-based MEDAs learn a Bayesian network (a directed acyclic graphical model, in which the nodes represent the variables and the arcs indicate the dependencies) as the probabilistic model for solution sampling \cite{ahn2007multiobjective, pelikan2005multiobjective}.}
\item{Regularity model-based MEDAs use a local PCA to build a probabilistic model based on the regularity property of MOPs, e.g., the PS of a continuous MOP is a piecewise continuous ($m$-1)-D manifold induced from the Karush-Kuhn-Tucker condition \cite{zhang2008rm, li2014improved, wang2012regularity, sun2018improved, sun2019new}.}
\item{Mixture distribution-based MEDAs generally divide the population of candidate solutions into several groups by a clustering method in the objective space, followed by learning a separate probabilistic model for each group to form a mixture distribution, which can improve the population's diversity \cite{bosman2002multi, karshenas2013multiobjective, karshenas2011multi}.}
\item{Inverse model-based MEDAs sample promising objective vectors in approximating the PF at first, followed by mapping them back to the search space \cite{farias2021moea,cheng2015multiobjective,guerrero2019multi}. Since the solutions set exhibiting a certain distribution in the search space does not necessarily mean that its mapping in the objective space also shows the same distribution, inverse modeling from the objective space to the search space is proposed to sample solutions.}
\end{enumerate}
\begin{figure*}[!t]
\centering
\includegraphics[width=6.6in]{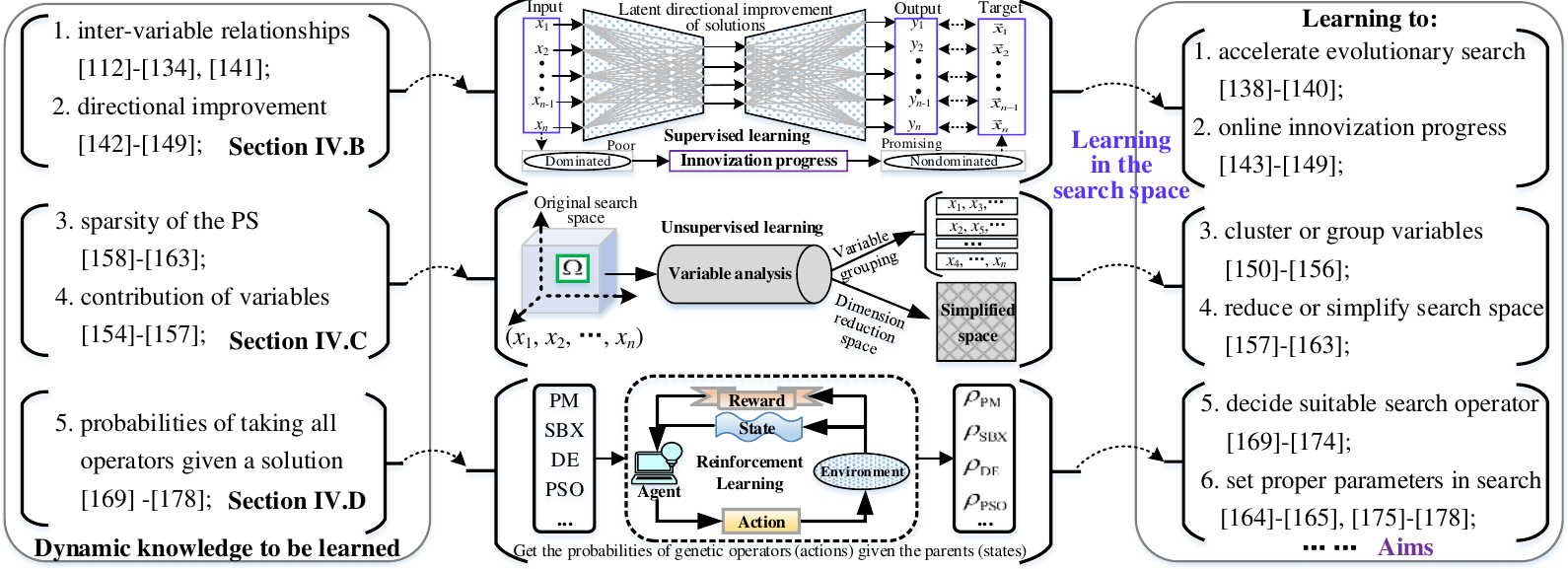}
\caption{Illustrating what knowledge can be learned in the search space via various ML models and the aims of learning this dynamic knowledge.}
\label{fig_sim}
\end{figure*}

Without involving genetic operators, these generators directly sample solutions in the search space by the learned distribution models. Then, these models are updated iteratively so that the resultant solution set can approach the PS. However, MEDAs are computationally expensive in building graphical-based models on LMOPs with a large-scale search space. Besides, they are likely to generate poor solutions when using the regularity-based models on the target MOP without apparent regularity \cite{sastry2005limits}. Probabilistic modeling may be a promising way to solve LMOPs, as studied in \cite{hong2022solving}. But it is only suitable for handling those LMOPs in which the mapping between the target space and search space is easy to simulate.

Recently, some newly developed MEDAs use generative adversarial net (GAN) to sample solutions in the large-scale search space for LMOPs. To train a GAN, the population is classified into a real set (with promising solutions) and a false set (with poorly performing solutions) \cite{he2020evolutionary,wang2021manifold}. Obviously, the performance of these adversarial learning-based generators strongly depends on the partitioning of the population. Some classification models, e.g., support vector machine (SVM), can be built to classify the population. In addition to learning probabilistic models to sample solutions, there have been a lot of studies on building ML models to assist the evolutionary search of generators, which are elaborated below.

\subsection{Learnable Generators via Supervised Learning}
Inspired by the idea that considering both randomness and opposition is better than pure randomness in neural networks, opposition-based learning strategies are used in some MOEAs to sample the initial population so that their evolutionary search starts from a relatively good state \cite{ma2014moea,ewees2021new}. To improve the search capability, competitive learning-based search strategies are developed in some MOEAs, in which the population is classified into a \textit{winner} set and a \textit{loser} set. Then, the search is guided in a way that the \textit{losers} move closer to the \textit{winners} \cite{tian2019efficient,liu2021comprehensive}. Since \textit{losers} are guided toward \textit{winners} by competitive learning, the search capability of the generator depends greatly on the quality of these \textit{winners}. Once they get into a dilemma (e.g., falling into local optima), the whole evolutionary population will converge slowly. More importantly, the \textit{winners} are often driven by genetic operators to have possibly slight improvement. How these better-quality \textit{winners} evolve further with a faster convergence speed becomes a challenging task.

To address this puzzle and accelerate the search speed, some efforts are proposed by designing online \textit{innovization} operators. Here, the newly coined \textit{innovization} is an abbreviation of ``innovation via optimization''. It is originally defined as a post-optimality analysis of the obtained optimal solutions \cite{deb2006innovization}. The offspring (reproduced via genetic operators) is further processed by \textit{innovization} operators to advance along with the learned directions of performance improvement \cite{mittal2021learning}. After finding a set of optimal solutions by an MOEA, \textit{innovization} is a task that uses various data-mining and ML techniques to automatedly unveil innovative and key design principles in these optimal solutions, such as inter-variable relationships, commonality principles among optimal solutions, and differences that make them mutually distinct \cite{bandaru2010automated}. Inspired by this, an \textit{innovization} study is proposed in \cite{deb2012hybrid} to learn salient search rules and clues for designing \textit{innovized} local search operators. They can make the population converge faster without consuming functional evaluation compared to traditional local search methods. In extensions of \textit{innovization}, the concept of knowledge-driven optimization is introduced in \cite{hitomi2017extracting}, in which MOEAs absorb the knowledge (e.g., the latent patterns shared by high-quality solutions) learned from intermediate solutions to guide their search away from mediocre solutions and toward potentially promising areas. This is an online learning process from deciphering what makes one solution optimal (or near optimal) in the final obtained solution set and revealing what makes one solution dominated by (or perform better than) another during the optimization process. Online \textit{innovization} aims to accelerate the evolutionary search and thus improve the efficiency of generators \cite{gaur2017effect}.

In an online \textit{innovization} operator, generally, a supervised learning model is first built and trained online. It expects to implicitly learn the historically directional improvements (e.g., from dominated to non-dominated) of solutions in the already explored search space. As shown in Fig. 5, a solution $\mathrm{x}=\left(x_1, x_2, \ldots, x_n\right)$ is identified to be performing poorly, while $\overrightarrow{\mathrm{x}}$ is of high quality. During the evolutionary process, a variety of $(\mathrm{x}, \overrightarrow{\mathrm{x}})$ solution pairs can be collected. Then, the adopted model, e.g., a multilayer perceptron \cite{liu2022learning} a random forest \cite{mittal2021enhanced}, or a deep neural network \cite{mittal2020learning,mittal2020ann}, can be trained using these labeled data. Here, $\mathrm{x}$ is the input and $\overrightarrow{\mathrm{x}}$ is its label, i.e., the expected target output. In this way, the trained model is believed to be able to capture the underlying pattern that reflects the directional improvement of solutions in the search space. Expectantly, a new offspring solution $\mathrm{x}^{\text {new }}$ generated by genetic operators can be repaired (or progressed) via inputting it into the well-trained model to get its improved version $\mathrm{y}^{\text {new }}$. This process of repairing offspring is called \textit{innovized} progress. It's possible to improve the search capability of generators in solving scaling-up MOPs due to its following four merits: 1) involving all conflicting objective information; 2) without requiring the cost of function evaluations; 3) the learned directional improvement of solutions is not fixed but is adaptive with the elapse of generations; 4) the moving from $\mathrm{x}^{\text {new }}$ to $\mathrm{y}^{\text {new }}$ by the learned model in the search space may result in a quantum leap in the objective space \cite{mittal2021enhanced}. Nonetheless, there are also four key considerations when customizing such an \textit{innovization} progress:

\textbf{Selection of the learning model:} It's flexible to choose potentially available supervised learning models to match the requirements. But, it's also reasonable to consider the actual computational cost caused by learning the selected model. For example, the random forest and multi-layer perceptron are respectively used in \cite{mittal2021enhanced} and \cite{liu2022learning} owing to their lightweight computational efficiency when compared with the deep neural networks adopted in \cite{mittal2020learning,mittal2020ann}.

\textbf{Collection of training data:} Paired training data, either from the previous or the current generation, are collected based on the performance of available solutions. Thus, when deciding to collect a pair $(\mathrm{x}, \overrightarrow{\mathrm{x}})$, it should consider that $\overrightarrow{\mathrm{x}}$ performs better than $\mathrm{x}$ (e.g.,  dominates $\mathrm{x}$ in a Pareto-based MOEA \cite{deb2006innovization}, the aggregation fitness of $\overrightarrow{\mathrm{x}}$ is better than $\mathrm{x}$ in a decomposition-based MOEA \cite{mittal2020ann}, and $\overrightarrow{\mathrm{x}}$ is most likely to guide $\mathrm{x}$ to converge rapidly \cite{liu2022learning}). Besides, the labeled population consisting of all $\overrightarrow{\mathrm{x}}$ solutions should perform better than the input population with all $\mathrm{x}$ solutions in terms of convergence-diversity tradeoff.

\textbf{Training of the adopted model:} Training a model itself is also an optimization problem, which involves many hyperparameters (e.g., model architecture, learning rate, and training epoch) that need to be set manually. Thus, many other issues may be encountered (e.g., overfitting). It is also worth exploring whether the model needs to be updated every generation (e.g., online or offline training).

\textbf{Advancement of the search capability with the learned model:} The expectation is that the search capability of the generator can be improved with the aid of this well-trained model. Specifically, the poor and mediocre solutions in the population can be repaired to converge fast in the learned promising direction. Meanwhile, high-quality solutions can be advanced to seek more diverse elite solutions. Nevertheless, it involves the following two considerations: 1) is it necessary to repair all the newly generated solutions? 2) is it necessary to run the \textit{innovization} progress in every generation?

Given above, the supervised learning methods are mainly used to improve the search capability of the evolutionary generator without simplifying the original search space.

\subsection{Learnable Generators via Unsupervised Learning}
As discussed in Section IV.A, the population is first divided into multiple clusters in some MEDAs by a clustering method. Then a probabilistic model is created on each cluster so that the created mixed model has a strong generalization ability in sampling solutions. Similarly, various clustering methods, e.g., $k$-means \cite{zhang2014clustering}, spectral clustering \cite{von2022clustering}, manifold clustering \cite{pan2019manifold}, and $k$-nearest neighbor (KNN) based graph \cite{wang2020adaptive}, are applied to assist evolutionary generators. To be specific, the crossover in a generator is restrictedly run only among solutions belonging to the same cluster. A plausible belief behind this is that mating similar solutions can create new solutions with higher quality and thus can improve search capability. Besides, different genetic operators can be implemented in different clusters instead of using the same search operators for the whole population, which can balance the exploration and exploitation of evolutionary search.

Additionally, clustering methods are used in \cite{zhang2016decision,ma2021adaptive} to classify the decision variables into multiple groups based on their contributions to the objectives. To be specific, all variables are divided into convergence-related and diversity-related groups based on the data collected by a variable contribution analysis. Afterward, different genetic operators are designed to search the subspace respectively consisting of two groups of variables, which can adaptively balance the convergence and diversity. Here, the clustering of variables requires analyzing the contribution of each variable (partial to convergence or diversity). Thus, this variable analysis process will consume a lot of function evaluations. Consequently, this kind of method often fails to do actual optimization of LMOPs when allocating a limited budget \cite{liu2021comprehensive}. Similarly, the decision variables are classified into highly and weakly robustness-related groups in \cite{du2018high} based on their contributions to the robustness of candidate solutions. Nonetheless, it is not enough to divide all variables into two groups because there are still many variables in each group for LMOPs. Therefore, the dimension of each group is further reduced by PCA in \cite{liu2020clustering}. In this way, a low-dimensional representative space can be obtained, and the generator can get a fast convergence speed by searching in this small-scale space.

The key question here is whether there is such a representative space for an LMOP. In this context, the concept of sparse LMOP arises \cite{tian2019evolutionary}, in which the values of most variables in optimal solutions are 0. The PS of a sparse LMOP can be approximated by finding a Pareto representation in a latent small-scale representative space. Since sparse LMOPs widely exist in real-world applications \cite{liu2022evolutionary} (e.g., neural network training, neural architecture search, community detection, signal processing, portfolio optimization, etc.), a variety of dimensionality reduction techniques (such as unsupervised neural networks \cite{tian2020solving}, data mining \cite{tian2020pattern}, logistic PCA \cite{ying2021multiobjective}, and network embedding \cite{liu2020multiobjective}) are applied to learn the representative space (also termed Pareto subspace) of sparse LMOPs. After that, new offspring can be generated by searching in this learned subspace, including the following three steps:
\begin{enumerate}
\item{\textbf{Encoding:} the solutions of an LMOP are encoded as small-scale representations;}
\item{\textbf{Searching:} the genetic operators are run on encoded representations to generate new representations;}
\item{\textbf{Decoding:} the newborn representations are mapped back to the original search space.}
\end{enumerate}
Consequently, the strategy of representation learning is proved to be effective in solving LMOPs with high sparsity.

Another promising way to improve the search capability is to adaptively select genetic operators or adaptively control/tune their parameters. Existing search strategies often need to set various hyper-parameters and exhibit quite different preferences in search patterns and scopes. Learning from historical search experience, some recent studies have been dedicated to the adaptive selection of better operators or the adaptive setting of appropriate parameters \cite{wang2020adaboost,zhang2007clustering}. For example, the AdaBoost-inspired multi-operator ensemble is proposed in \cite{wang2020adaboost} to define the credit assignment scheme for adaptively rewarding the involved operators. The clustering-based fuzzy system is customized in \cite{zhang2007clustering} to adaptively tune the crossover and mutation probabilities. The key component here is the credit assignment scheme used to assess the performance improvement brought by each participant operator \cite{hitomi2016classification}. Then, the higher-performing operator is activated by the generator to reproduce offspring in the next generation.

\subsection{Learnable Generators via Reinforcement Learning}
Adaptive operator selection or adaptive parameter setting are controllers in improving the search capability of evolutionary generators \cite{eiben2007reinforcement}. However, relying only on historical rewards, they still suffer from the dilemma of exploration versus exploitation. Accountably, the operators may be inaccurately punished due to the fact that one who performs poorly in previous generations may be effective in future generations. Besides, searching with poorly assessed operators according to their historical rewards may be beneficial for the solutions to escape from the local optimum \cite{tian2022deep}. To alleviate these issues, RL-assisted control strategies are proposed in \cite{wang2021cooperative,zhao2020online,consoli2016dynamic,zhao2020decomposition,li2019differential,huang2020fitness} and \cite{sakurai2010method,sun2021learning,kaur2020reinforcement,bora2019multi}, respectively, for operator selection and parameter setting. Compared with the adaptive control strategies, they have the following advantages:
\begin{enumerate}
\item{RL-assisted strategies rely on future rewards rather than historical rewards, which expect the selected operators to be effective in future generations rather than in the previous generations;}
\item{RL-assisted strategies can provide a more delicate decision on operator selection as they consider the characteristics of each solution. Besides, they can assign different operators for different parents rather than reward operators equally for all solutions.}
\end{enumerate}
Since the future rewards are unknown during the search process, it is needed to be deduced by training a relatively complex RL model. It includes the following four core components to be determined:

\textbf{Actions:} a set of genetic operators constitutes the discrete action space. Here, the same operator with different parameter settings can be considered as a different operator.

\textbf{States:} the decision variables of solutions defined in the search space (i.e., the environment) are regarded as states, meaning that each solution has its own state.

\textbf{Policy network:} Inputting a state, a policy network should output the probability of taking each action so that the agent can decide a reasonable next action. Generally, the probability of a decision increases for high reward values. To train the policy network steadily, the reward for a state-action pair should be set reasonably. It should take both convergence and diversity into consideration in the performance improvement obtained by newly generated offspring.

\textbf{Rewards:} Since the action space is discrete, the classical Q-learning can be used to approximate the action-value function \cite{tian2022deep}, which is always defined based on the discount sum of future rewards. Here, the reward is calculated based on the performance improvement of offspring obtained after taking a genetic operator. The reward corresponding to these actions is stored in a Q-table, where each grid represents the expected cumulative reward of an action given a state.

In short, the agent selects a long-sighted operator (i.e., action) to the MOEA according to the output of the policy network. Then, an offspring (i.e., a new state) is generated by using the selected operator to search in the variable space (i.e., environment). After that, the new state and performance improvement (i.e., reward) are sent to the agent for updating the policy network. With existing updated rules for Q-learning in place, the core of designing RL-assisted control strategies shifts to assessing the performance improvement from parents to offspring brought by the conducted genetic operator. Thus, the discriminators introduced in Section III can be used here to get a reasonable assessment of the performance improvement.

\section{Learnable Evolutionary Evaluators}
For MOPs with clear expression, the MOEA can directly use its evaluator to calculate the objective vector for each solution x, i.e., the given function vector F(x). However, the objectives of many practical MOPs often lack a closed form and may be expensive to evaluate from simulation data, which makes MOEAs unable to get the near-optimal solution set within a limited computing time \cite{li2022evolutionary}. To address this issue, learnable evaluators use surrogate models to simulate the objective functions or fitness of solutions. Correspondingly, learnable evaluators can be roughly divided into two different types: learnable function evaluators \cite{knowles2006parego} and learnable fitness evaluators \cite{pan2018classification}. The details of them are reviewed below.

\subsection{Learnable Function Evaluators}
For learnable function evaluators, the calculations of real expensive objective functions are simulated by directly training a cheap regression-based surrogate model, e.g., Kriging model \cite{zhang2009expensive,chugh2016surrogate,song2021kriging}, dropout neural network \cite{guo2021evolutionary}, radial basis function \cite{lin2022adaptive,guo2018heterogeneous}, hybrid or ensemble models \cite{habib2019multiple,lin2021ensemble}. In this way, to save computing time in each generation during evolution, only a small number of promising solutions (sampled by the surrogates) are eligible for real function evaluations. Then, they are used to update the surrogates, which are applied to evaluate the remaining solutions functionally. Thus, the trained inexpensive surrogates often need to be accompanied by the model management to sample promising solutions. Here, various infill sampling criteria are proposed for model management \cite{zhan2017expected}, such as lower confidence bound, probability improvement, expected improvement, and multiobjective infill criterion \cite{tian2018multiobjective}. Several survey papers have been proposed to detail surrogate-assisted EAs. Interested readers should refer to these related reviews \cite{jin2019data,li2022evolutionary,jin2005comprehensive}.

\subsection{Learnable Fitness Evaluators}
In addition to directly simulating real function evaluations, some learnable fitness evaluators switch to building surrogates for approximating scalar aggregation fitness of multiple objectives in decomposition-based MOEAs \cite{liau2013machine,rahat2017alternative} or for simulating a performance indicator (like hypervolume) in indicator-based MOEAs \cite{ponweiser2008multiobjective,picheny2015multiobjective}. Besides, the evaluation of objective functions may be less important in solving an MOP if the dominant relationship between solutions is known. Inspired by this, some learnable fitness evaluators expect to skip the function evaluator, and directly learn a classifier to determine the quality or dominance levels of solutions \cite{seah2012pareto}. Ideally, it is possible to learn a binary classifier to determine which solutions should survive into the next generation (positive class) and which should not (negative class) \cite{hao2020binary}. Thus, these classification-based evaluators, also acting as discriminators, can concurrently balance the convergence and diversity of population if an appropriate classification boundary is defined \cite{chugh2019survey}. To learn Pareto-dominance relationship between solutions, ten off-the-shelf classification models are studied in \cite{bandaru2014performance}. Deep feedforward neural networks are used to predict the $\theta$-dominance levels of candidate solutions in \cite{yuan2021expensive} and to learn Pareto-dominance relations between solutions and references in \cite{hao2022expensive}. KNN model is utilized in \cite{zhang2018preselection} for filtering candidate solutions and the support vector machine is employed in \cite{lin2016decomposition,sonoda2022multiple} to classify solutions based on each subproblem's scalarization fitness.

\section{Learnable Transfer Evolutionary Modules}
Taking the evolutionary process of solving an MOP as an optimization exercise (or task), the aforementioned MOEAs generally show the following three characteristics: 1) they often start an evolutionary search from scratch with a random population; 2) they assume that all MOPs are mutually independent and thus solve each MOP as a separate task; 3) their problem-solving ability does not grow with the optimization exercises they have undergone. However, such an optimization manner of solving each MOP independently is inconsistent with the problem-solving behaviors of human beings \cite{gupta2017insights}. A prospective MOEA should be able to improve its problem-solving ability by accumulating experience from previous optimization exercises. Then, it can solve relevant MOPs at hand in a more efficient way. Taking this cue, research into reusing transferable knowledge (from the experiences of optimizing source MOPs) to assist in the optimization of target MOPs has attracted extensive attention. This kind of learnable MOEAs is always coupled with transfer learning \cite{tan2021evolutionary}. They expect to find such a shortcut that the evolutionary population could 1) have a higher starting state than starting from scratch; 2) have a faster convergence rate; and 3) eventually be closer to the target PF or PS. This section will sort out the efforts in improving the performance of three evolutionary modules (i.e., generators, discriminators, and evaluators) through transfer learning. The details are provided and discussed below.
\begin{figure}[!t]
\centering
\includegraphics[width=3.2in]{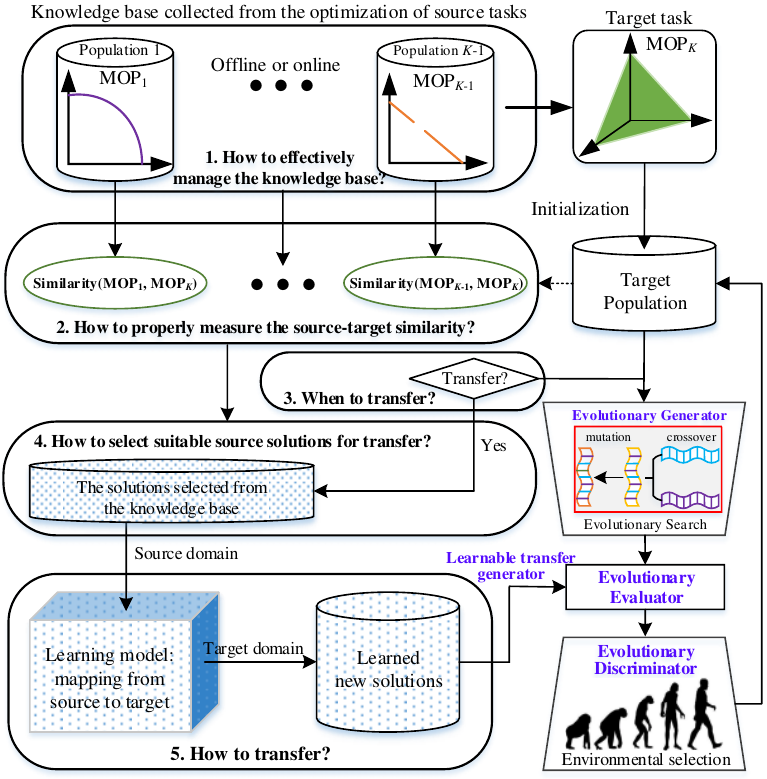}
\caption{The basic framework of solution-based transfer generators and the main challenges when designing this kind of transfer generators.}
\label{fig_sim}
\end{figure}
\subsection{Learnable Transfer Generators}
A learnable transfer generator continually updates and enriches its search experience by learning from the optimization of source MOPs \cite{liu2022DRN}. Then, its search efficiency can be improved with the assistance of transferred knowledge. Existing efforts on learnable transfer generators mainly regard solutions of source MOPs as knowledge to be transferred. As shown in Fig. 6, solution-based transfer generators are primarily studied by addressing the five key challenges, as follows:

\textbf{How to effectively manage the knowledge base?} The discovered optimal solution sets in solving source MOPs are collected for updating the knowledge base. Here, a threshold is always preset to determine the number of saved solutions for each source MOP. Once the cardinality of a source MOP in the knowledge base exceeds the threshold, some related solutions need to be filtered out. For example, the source task completed later is believed to be more closely related to the target task in solving a dynamic MOP \cite{xu2021online}. Thus, the optimal populations of adjacent source tasks are preferred to be saved when pruning the knowledge base \cite{raquel2013dynamic}. For periodic dynamic MOPs, the source tasks at the same cycle stage as the target task are preferred by the knowledge base \cite{chen2019novel}. It is noteworthy that dynamic optimization is regarded as a special type of sequential optimization as shown in Fig. 2. For more information on dynamic optimization, please refer to this latest survey \cite{jiang2022evolutionary}. When solving multiple MOPs simultaneously from scratch, the optimization of each MOP is both a target task and a source task \cite{liu2022DRN}, in which the knowledge base always consists of the current populations they explored.

\textbf{How to properly measure the source-target similarity?} The negative transfer may occur when the PS of the source MOP is unrelated or even opposite to that of the target MOP. To alleviate this issue, it is essential to exactly measure the source-target similarity based on the distribution distance between their obtained solutions. Various measures of distributional distances, e.g., mutual information \cite{fu2017new}, Wasserstein distance \cite{zhang2019multisource}, Kullback-Leibler divergence \cite{da2018curbing}, and maximum mean discrepancy \cite{jiang2017transfer}, have been used to quantify the similarity between source-target MOPs. Since the obtained solutions change with the evolutionary process, the source-target similarity should be always updated online, which is often intractable. For example, change detection in dynamic optimization is implicitly a process to measure whether two adjacent tasks are similar or not. Most of the related works learn a prediction model to detect environmental change \cite{hu2021handling}. However, the solution sets collected in different environments for prediction do not always follow the same distribution \cite{ma2021feature}, which will mislead the search. In heterogeneous source-target scenarios, it is often necessary to do padding or map all the solutions to a common space \cite{da2018curbing}. In multitasking scenarios, the online population may only cover a subregion of the search space, which further increases the difficulty of online similarity measurement. Thus, the inter-task gene similarity is defined in \cite{ma2021improving} by a probabilistic model to feature each gene.

\textbf{When to transfer?} Too frequent knowledge transfer needs to spend a huge amount of function evaluation on the transferred solutions, which is not conducive to efficient optimization \cite{ruan2019and}. It will also lead to a large increase in the probability of negative transfer. Besides, solutions transferred to the target MOP may be only useful at certain evolutionary stages. Thereby, it is worth discussing whether it is necessary to transfer knowledge from source and when is appropriate to transfer. For example, knowledge transfer is generally put forward only when the environment change is detected in dynamic optimization \cite{jiang2016steady}.

\textbf{How to select suitable source solutions for transfer?} The types of source tasks will be diverse with the increase in optimization experiences.  Accordingly, the selection of appropriate source tasks for transfer learning becomes critical. The most intuitive way is to select the source task with the highest similarity to the target task. However, due to the potential randomness of evolutionary search, the population's distribution is constantly changing during the search process. Thus, the current source-target similarity hardly reflects their actual relatedness. In fact, without a \textit{priori} knowledge about source and target MOPs, it is difficult to correctly judge the difference between source and target from their obtained solutions \cite{yazdani2019scaling}. Nevertheless, the source-target similarity suggests some latent connection between them. A variety of strategies for selecting source tasks based on the source-target similarity are presented in \cite{zhang2019multisource}, including the nearest selection, weighted selection, top-$k$ selection, and their mixed version. In \cite{xue2021evolutionary}, according to the contribution of variables, the source tasks saved in the knowledge base are divided into diversity-related and convergence-related source tasks. In \cite{liang2021evolutionary}, multiple highly similar source tasks are first selected for each target task, which will learn from these multiple sources using a local distribution estimation.

In addition, all the solutions of source tasks are potential transferable knowledge, some of which are helpful while others may be futile or even counter-productive. Hence, the selection of solutions with high-quality knowledge is also an important foundation to underpin positive transfer. In \cite{lin2019multiobjective}, each solution will get a class label (''+'' or ''$-$'') via an incrementally trained naive Bayes classifier, and then only the solutions whose predictive class labels are ''+'' can be the candidate transferable solutions. In \cite{lin2020effective}, the transferred solution will be marked as positive once it is non-dominated for the target task, and then only neighbors of positive transferred solutions can be the candidates for future knowledge transfer. The concept of ''transfer rank'' is defined in \cite{chen2022multi} to further quantify the priority of each historical positive transferred solution, in which those solutions with higher ranks are preferred to be transferred. In \cite{han2021self}, a knowledge estimation metric is defined to imply the effectiveness of solutions. In \cite{wang2022multiobjective}, each subproblem of the target task maintains an external neighborhood with the subproblems of source tasks via grey relation analysis. Then, the neighborhood information acts as a bridge to limit the selection of transferable solutions.

\textbf{How to transfer?} In homogeneous source-target scenarios, the solutions from the source domain can be directly regarded as a new solution in the target domain, such as in most dynamic optimization cases \cite{jiang2022evolutionary}. However, when facing heterogeneous source-target scenarios, a suitable transfer model is required to be learned for mapping solutions from the source domain to the target domain. For example, a source solution $\mathrm{y}$ is mapped into the target domain by an autoencoder model, which can generate a new solution $\mathrm{x}$ with the linear mapping $f: \mathrm{x}=\mathrm{y} M$ \cite{feng2017autoencoding,feng2015memes,feng2020solving}. Here, the matrix $\mathrm{M}=\left(\mathrm{QP}^{\mathrm{T}}\right)\left(\mathrm{PP}^{\mathrm{T}}\right)^{-1}$ is learned by minimizing $\|\mathrm{QM}-\mathrm{P}\|^2$, where $\mathrm{Q}$ and $\mathrm{P}$ respectively represent the populations for source and target MOPs. For achieving nonlinear mapping, the kernelized autoencoder is developed in \cite{zhou2021learnable} and \cite{zhou2021evolutionary} for solving sequential MOPs and dynamic MOPs, respectively.

Moreover, a robust transfer model should be built by considering the distribution shift between source and target tasks. Domain adaptation methods are designed to overcome this shift, which can reduce the discrepancy between source and target tasks by learning their domain-invariant feature representations. Specifically, an intermediate subspace is constructed to find a learnable alignment matrix that transforms a source solution into the target one. As listed in Fig. 7, the subspace can be composed of several most significant eigenvectors (representing the necessary features) induced by PCA, as studied in \cite{liang2020evolutionary} and \cite{tang2020regularized}. A very similar method that treats the search space as a manifold is proposed in \cite{chen2020learning,jiang2020fast}. Then, all search spaces modeled as a joint manifold are projected into a latent subspace by solving a generalized eigenvalue decomposition problem. The coordinate system is defined in \cite{tang2021multifactorial} to identify some common modality in a proper intermediate subspace, which is generated along the geodesic flow starting from the source subspace to the target subspace. Furthermore, domain adaptation based on probability distribution is also widely used to sample transferable solutions, e.g., the transfer component analysis used in \cite{jiang2017transfer}, the joint distribution adaptation studied in \cite{ma2021feature}, and the moment distribution matching and representation learning of source-target are respectively proposed in \cite{xue2021evolutionary} and \cite{lim2021solution}.

Most efforts on learnable transfer generators focus on reusing solution-based knowledge. However, in practice, the representations of knowledge should be diverse, such as the trend of environmental change in dynamic optimization, the search behaviors of different genetic operators, and the latent attributes that affect evolution \cite{zou2019dynamic}. Thus, the search trend is learned in \cite{liang2021multiobjective} by a maximum likelihood estimation model, adaptively controlling the evolutionary search (e.g., search step size and range). A neural network and a SVM are respectively trained in \cite{liu2019neural} and \cite{cao2019evolutionary} to learn the transfer trend of environment changes for reproducing promising solutions in new environments. In \cite{li2021meta}, the concept of ''meta-knowledge'' is introduced to evolve the task-specific knowledge.
\begin{figure}[!t]
\centering
\includegraphics[width=3.3in]{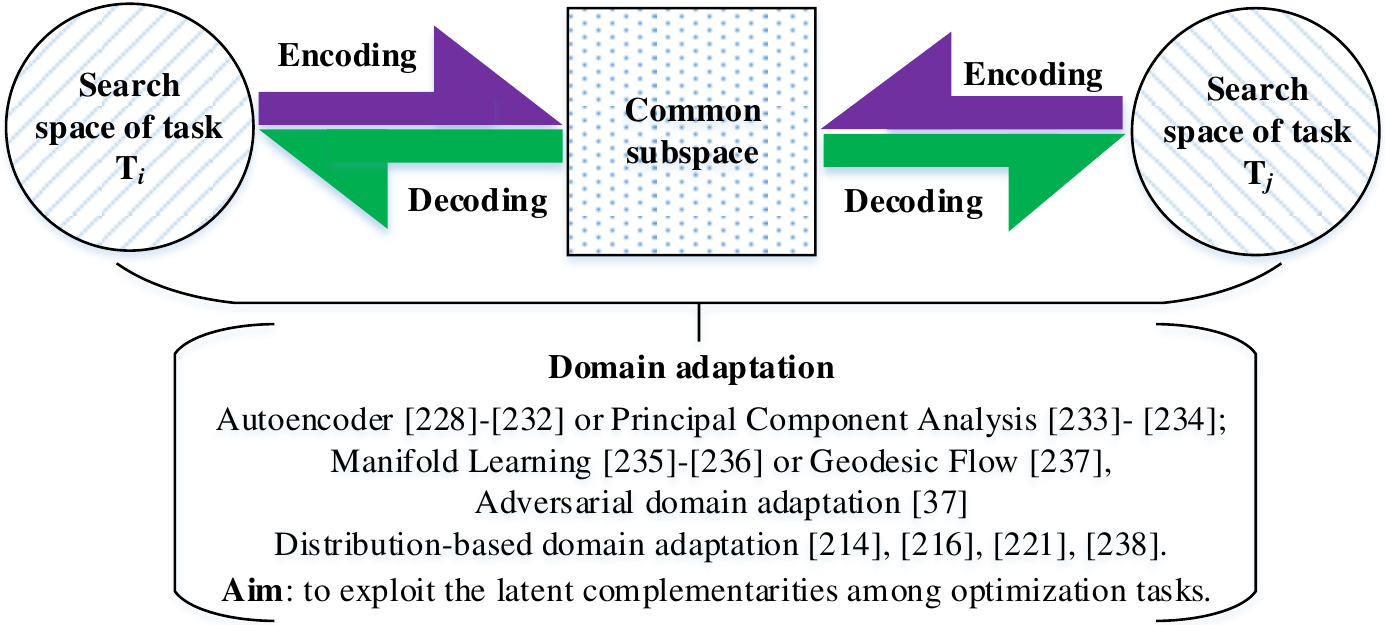}
\caption{The basic framework of solution-based transfer generators and the main challenges when designing this kind of transfer generators.}
\label{fig_sim}
\end{figure}
\subsection{Learnable Transfer Evaluators and Discriminators}
Existing efforts on evolutionary transfer optimization mainly focus on improving the search capability of generators via solution-based knowledge transfer. There are still few efforts on learnable transfer evaluators and discriminators for EMO. This subsection will summarize the related studies. Among the basic challenges, how to alleviate negative transfer and ensure maximum positive transfer, even in scenarios of source-target domain mismatch, is also the focus of this research area \cite{lim2021non}.

Learnable transfer evaluators are mainly designed to handle EMOPs. Multiple surrogates are built respectively for different computationally expensive tasks, which are jointly solved to reduce the number of function evaluations \cite{liao2020multi}. Intuitively, transferable knowledge embedded in cheap surrogates can be reused to avoid computationally expensive surrogate modeling \cite{li2022overview}. For example, the evaluation of different objectives may involve distinct computer simulations with different computational complexities. To evaluate these non-uniform objectives, different surrogates should be created. In this scenario, the useful knowledge (e.g., parameters) from the surrogates for cheap (or fast) objectives can be transferred to assist the surrogates for expensive (or slow) objectives \cite{wang2020transfer,wang2022transfer}. Besides, auxiliary (or local) surrogates trained on subsampled data (or local data) can be regarded as helpful knowledge to assist the learning of global surrogates on all available data (or large data) \cite{ding2017generalized,chen2022scaling}. For solving EMOPs in dynamic environments, the previously archived data is transferred to the new environment to jump-start (or warm start) its surrogate modeling process \cite{li2022data,fan2020surrogate}. In \cite{min2017multiproblem}, the knowledge learned in distinct surrogate models is transferred to augment the optimization of new possible related target tasks. To avoid the complex covariance evaluation in Bayesian optimization, multitask conditional neural process nets are built as surrogate models to replace the Gaussian process. The proposed surrogate models augment the observed dataset with several related tasks to estimate model parameters confidently \cite{luo2020novel}.

The evaluators discussed above can be regarded as a discriminator in certain cases, such as when there is only one objective function per task. In addition, a discriminator based on transfer learning between different clusters is proposed in \cite{li2022reducing}, which aims to distinguish the quality of solutions in the new environment. Ensemble prediction models are learned in \cite{guo2019ensemble} to find an acceptable but more robust PF during environmental changes. An RL model is built in \cite{zou2021reinforcement} to respond to the movement and different severity degrees of the Pareto front in time-varying environments. For EMOPs and DMOPs with large-scale search spaces, it may be a promising research direction by transferable inverse modeling \cite{zhou2013population,zhang2021inverse,gee2016solving}. In \cite{wei2021towards}, a multi-step nonlinear regression is designed to quantify the convergence status and estimate the intensity of knowledge transfer, aiming at achieving adaptive allocation of computational resources in multitasking scenarios.

Finally, an MOEA can collect a large amount of diverse data and optimization knowledge after solving a series of different MOPs, but how to reuse such data and knowledge smartly in promoting the future optimization task is still in its infancy \cite{fang2023domain}. The intention of this section is not to review the existing work on multitasking, expensive, dynamic optimization, etc., but to give an investigation of knowledge transfer in these different optimization scenarios \cite{guo2022knowledge}. Interested researchers can refer to their latest surveys \cite{wei2021review,osaba2022evolutionary,li2022evolutionary,jiang2022evolutionary} for getting insight into more details. Here, we suggest a recent short letter \cite{osaba2022emultitask}, which summarizes the fundamental research questions, limitations, challenges, and directions for the future related to evolutionary transfer or multitask optimization.

\section{Challenges and Future Opportunities}
Despite the efforts of MOEAs powered by ML methods, the potential of learnable MOEAs for scaling-up MOPs has not yet been fully explored. These efforts are only the first steps and still have much room for improvement \cite{zhan2022learning}. Here, we discuss the challenges in advancing learnable MOEAs and provide insights on the prospects of future opportunities.
\subsection{Challenges}
At first, both MOEAs and ML methods are insufficiently studied in theory. The theoretically fundamental studies of learnable MOEAs are challenging. It may be difficult to mathematically define the diverse types of knowledge learned during evolution. It's also hard to quantify how much the learned knowledge can help MOEAs in solving various scaling-up MOPs. Furthermore, the theoretical convergence and computational complexity analysis of learnable MOEAs may be more difficult to advance \cite{chai2013evolutionary}.

Secondly, the understandability and robustness of learnable MOEAs for MOPs are challenging. MOEAs often generate changeable and locally distributed data, and ML is an optimization task of function approximation from given data. Thus, it may be challenging to learn the expected function (e.g., what we want the ML model to do). Consequently, the approximated function of the adopted ML model may be noisy and even contain contradictions to the target optimization of MOPs. Therefore, there is still a long way to better understand exactly what ML models can contribute to evolutionary optimization tasks and to clearly explain why the adopted ML model works well on specific optimization tasks.

Thirdly, developing efficient learnable MOEAs for various scaling-up MOPs remains an open challenge. ML models are usually susceptible and computationally expensive. There are so many existing ML models and they are still being updated rapidly. To efficiently solve the scaling-up MOPs of interest, it is difficult to determine what kind of ML models should be assembled into MOEAs. It is rarely wise that research studies provide a tiny accuracy benefit at the cost of massive increases in model complexity. Even the addition of one or two seemingly innocuous data dependencies can slow down the progress.

Fourthly, great difficulties remain regarding the fairness and verifiability of performance comparison for newly developed learnable MOEAs. As explained in \cite{ishibuchi2022difficulties}, it is difficult to guarantee absolute fairness in performance comparison of MOEAs, which involves various factors to be considered. Specifically, different population sizes, test problems, performance indicators, or termination conditions may result in different performance comparison results. Besides, most of the existing ML models are inherently difficult to explain in terms of performance. Current studies verify and analyze the performance of a new learnable MOEA in an intuitive way, which is not beneficial for the development of research in this field. Thus, it is of great significance to take this aspect into consideration to endow learnable MOEAs with scientific rigor and transparency \cite{osaba2021tutorial}.

Last but not the least, there is still a long way to go to apply learnable MOEAs for tackling real-world MOPs. The performance of most learnable MOEAs is validated only by synthetic MOPs. Whether the real challenge of solving these practical MOPs is the same as the synthetic MOPs is a very meaningful but challenging inquiry.

\subsection{Future Opportunities}
In order to drive the development of learnable MOEAs for solving scaling-up MOPs, several potential directions and opportunities for future study are provided below.

One direction is to improve the effectiveness and efficiency of learnable MOEAs. With this goal, we should prefer to couple an ML model with high performance and low computational cost into MOEAs, rather than deliberately copying the popular ML models. The high performance here should depend on whether the model can exactly contribute to improving the problem-solving ability. The low computational cost should not just be about the function evaluation cost in training an ML model. Learning the model is also an optimization process, so determining how to achieve a better balance between learning and optimization is a meaningful topic for future study. For example, design lightweight classifiers to replace the regression-based function evaluations for expensive MOPs \cite{zhang2022dual} and combine some local search methods to accelerate the population's convergence speed \cite{li2022surrogate}.

Designing learnable MOEAs to solve more complex or special MOPs is also a promising research direction, e.g., surrogate-based Q-learning for dynamic time-linkage EMO \cite{zhang2022surrogate}, surrogate models for high-dimensional expensive EMO \cite{liu2022reference}, adaptive surrogates for expensive constrained multimodal EMO \cite{zhang2022objective}, evolutionary transfer for multi-scenario EMO \cite{jiang2022multi}, multitasking under interval uncertainties \cite{yi2020multifactorial}, evolutionary multitasking for constrained EMO \cite{qiao2022evolutionary,ming2022constrained}, federated learning for secure EMO \cite{liu2022secure}. Besides, we should not be satisfied with effectively solving synthetic MOPs, but pursue more smart or intelligent optimizers for practical applications, e.g., multitasking and hyper-heuristic are unified in \cite{hao2020unified} with a graph-based framework \cite{yan2019graph}, hybridization of MOEA and RL for multiobjective orienteering optimization \cite{liu2022hybridization}. Thus, designing various scaling-up MOP benchmarks closely related to practical applications is another meaningful topic that can further advance this field \cite{bartz2020benchmarking}, e.g., data-driven MOPs \cite{liu2022performance}, benchmarking neural architecture search as MOPs \cite{lu2023neural}, and benchmarking feature selection as MOPs \cite{nguyen2022constrained}.

Advances in hardware and software (e.g., GPU and parallel computing \cite{tan2015survey}), facilitating larger models and faster training, have greatly promoted the research of deep learning. Similar technological advances for learnable MOEAs could massively improve their problem-solving efficacy \cite{li2020generation}. If possible, bespoke hardware and software could be developed to run learnable MOEAs for tackling the challenges of ever scaling-up MOPs we may encounter in the future. For example, the GPU-based knowledge transfer mechanisms are proposed in \cite{huang2021toward} to handle many-task scenarios and the batch recommendation approach is presented in \cite{li2022batched} to evaluate multiple samples parallelly for data-driven optimization \cite{wang2018random}. Finally, the automated design of learnable MOEAs is also a promising development direction \cite{yi2022automated}.

\section{Conclusions}
When solving the challenging scaling-up MOPs, designing effective learnable MOEAs has become a hot topic in evolutionary multiobjective optimization due to their competitive performance and tremendous potential compared with traditional model-free MOEAs. Equipped with various machine learning techniques, different types of learnable MOEAs, as summarized in this survey, have been proposed in recent years. These learnable optimizers show promising problem-solving ability in responding to the grand challenges from a wide scale of complex MOP scenarios, e.g., MaOPs, LMOPs, EMOPs, DMOPs, SMOPs, MMOPs, etc. To clarify the underlying collaboration of applying ML methods to assist MOEAs, this paper defines the learnability of MOEA from four different learnable evolutionary modules, i.e., discriminator, generator, evaluator, and transfer modules. Then, a systematic review is provided for the advancement of each direction. Finally, the challenges of developing learnable MOEAs are provided and several future opportunities are presented to drive the development of learnable MOEAs for scaling-up MOPs.

\def\bibfont{\fontsize{8.0}{9.8}\selectfont}

\begin{IEEEbiography}[{\includegraphics[width=1in,height=1.25in,clip,keepaspectratio]{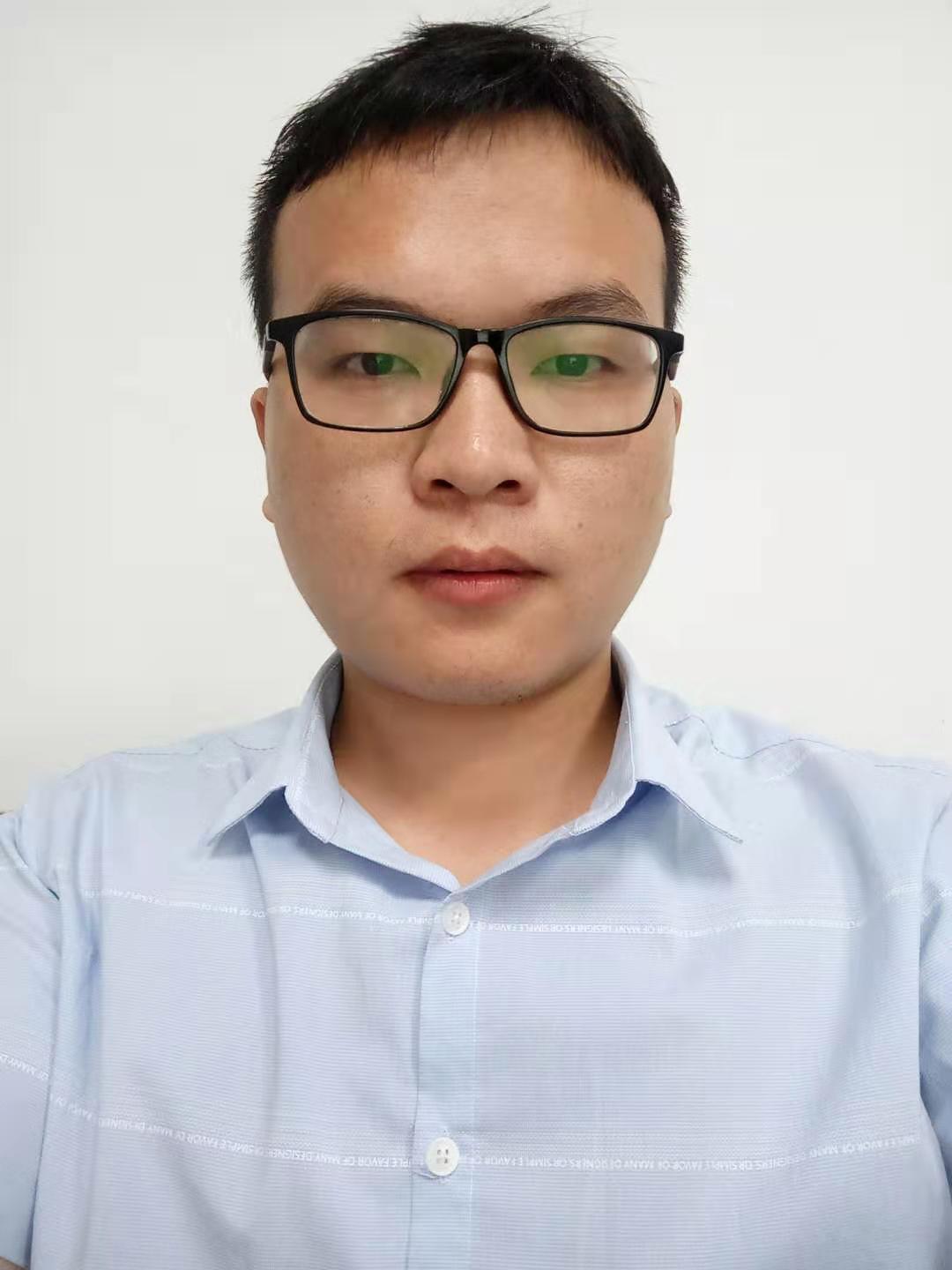}}]{Songbai Liu}
(Member IEEE) received the B.S. degree from Changsha University and the M.S. degree from Shenzhen University, China, in 2012 and 2018, respectively. He received the Ph.D. degree from Department of Computer Sciences, City University of Hong Kong, in 2022.

He is currently an assistant professor in College of Computer Science and Software Engineering, Shenzhen University. His research interests include evolutionary algorithms + machine learning, evolutionary large-scale optimization, and their applications.\end{IEEEbiography}

\begin{IEEEbiography}[{\includegraphics[width=1in,height=1.25in,clip,keepaspectratio]{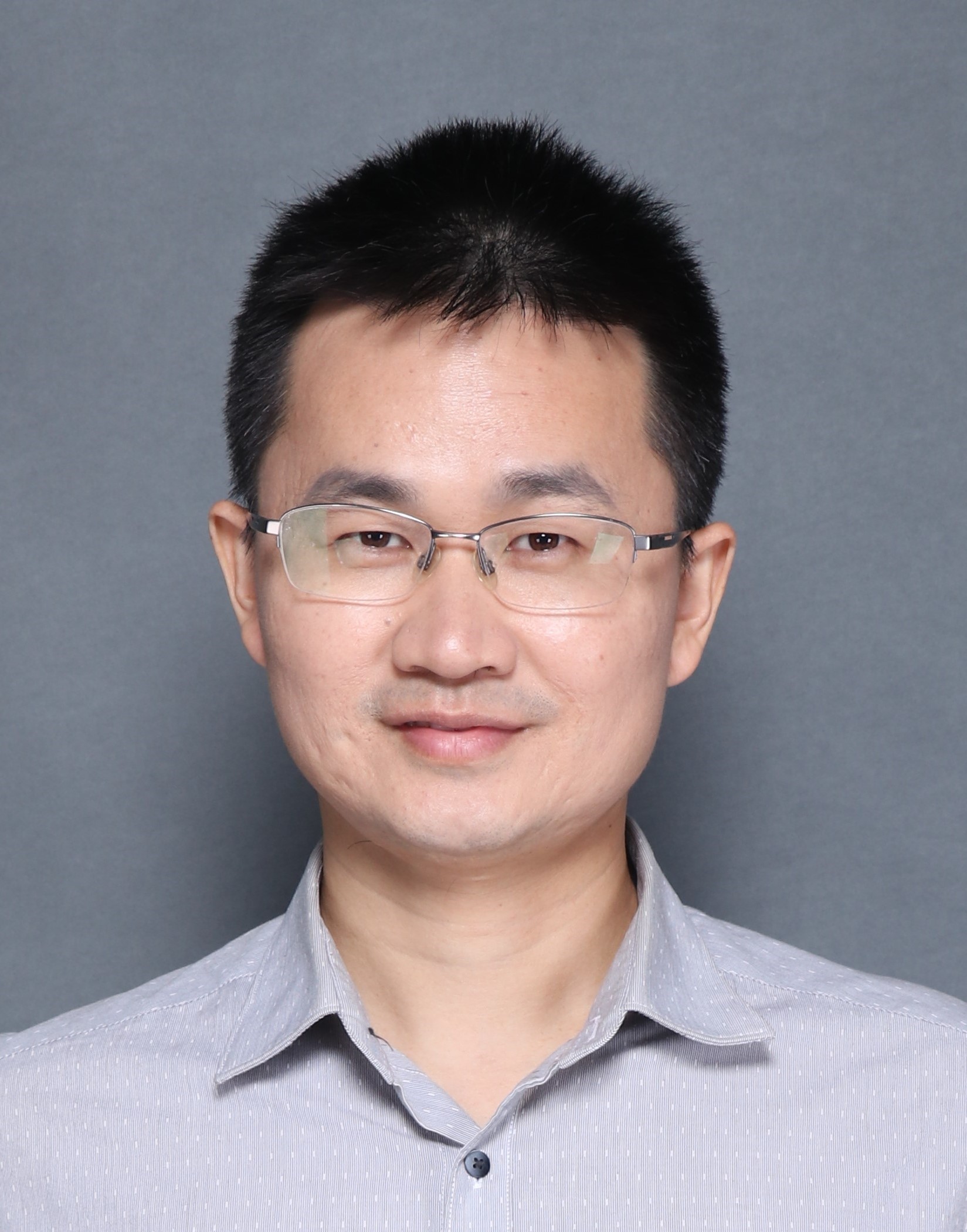}}]{Qiuzhen Lin}
(Member IEEE) received the B.S. degree from Zhaoqing University and the M.S. degree from Shenzhen University, China, in 2007 and 2010, respectively. He received the Ph.D. degree from Department of Elec-tronic Engineering, City University of Hong Kong, Kowloon, Hong Kong, in 2014.

He is currently an associate professor in College of Computer Science and Software Engineering, Shenzhen University. He has published over sixty research papers since 2008. He is an Associate Editor of the IEEE TRANSACTIONS ON EVOLUTIONARY COMPUTATION and the IEEE TRANSACTIONS ON EMERGING TOPICS IN COMPUTATIONAL INTELLIGENCE. His current research interests include artificial immune system, multi-objective optimization, and dynamic system.\end{IEEEbiography}

\begin{IEEEbiography}[{\includegraphics[width=1in,height=1.25in,clip,keepaspectratio]{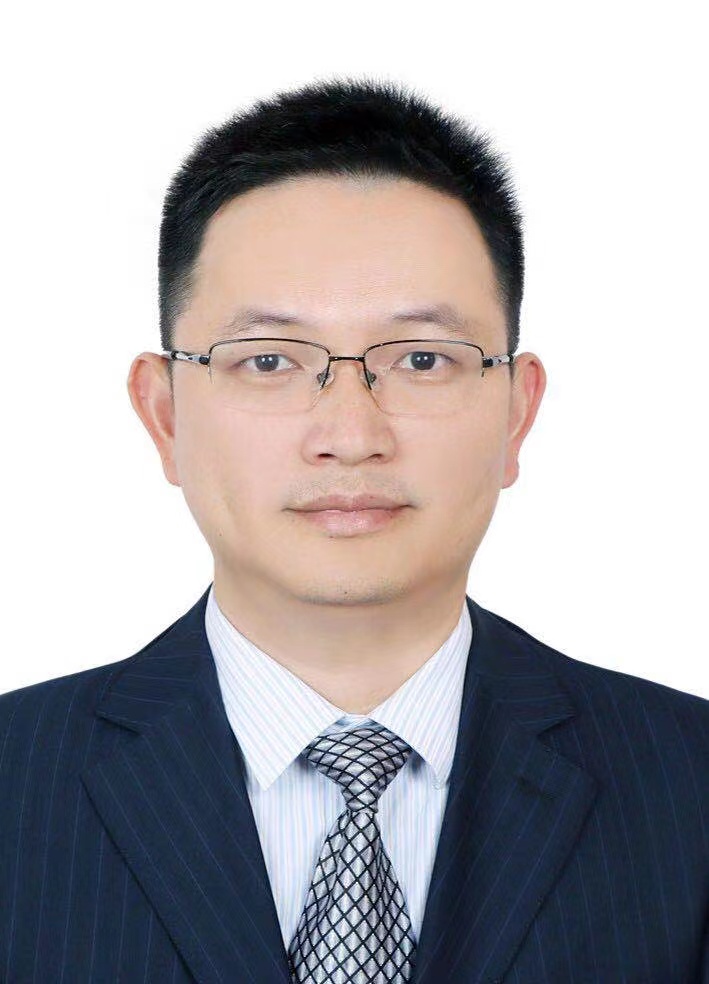}}]{Jianqiang Li}
(Member IEEE) received his B.S. and Ph.D. Degree in automation major from South China University of Technology, Guangzhou, China, in 2003 and 2008, respectively.

He is a professor at the College of Computer and Software Engineering of Shenzhen University. He led five projects of the National Natural Science Foundation and four projects of the Natural Science Foundation of Guangdong Province, China. His current research interests include robotics, embedded systems, and Internet of Things.\end{IEEEbiography}

\begin{IEEEbiography}[{\includegraphics[width=1.05in,height=1.25in,clip,keepaspectratio]{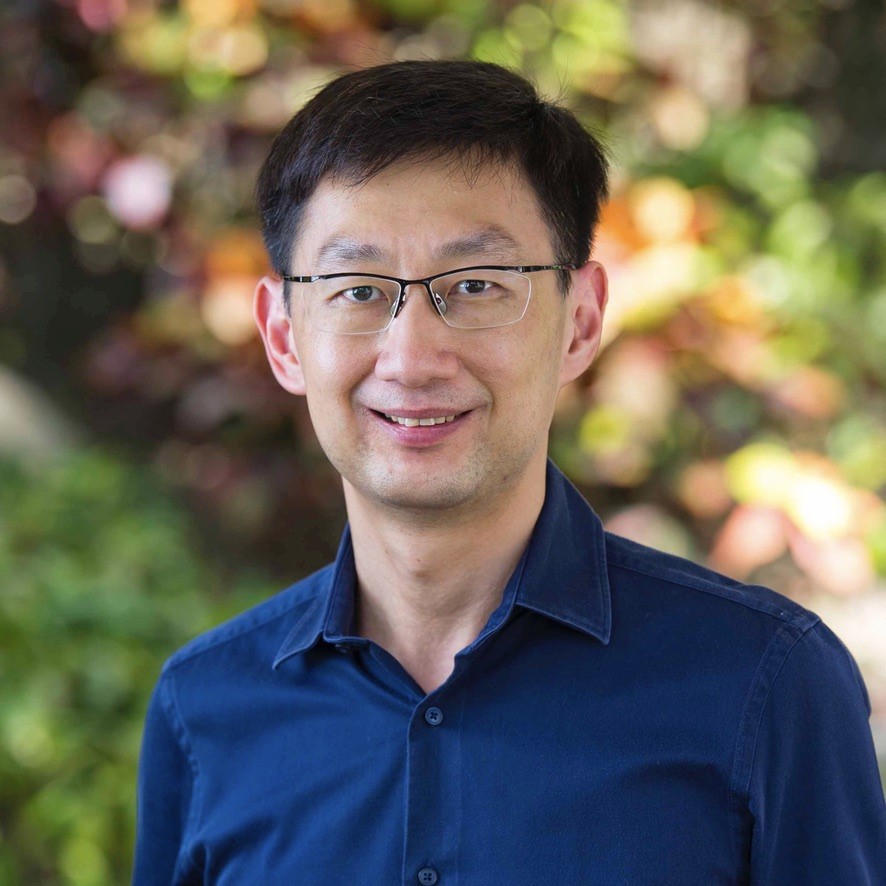}}]{Kay Chen Tan}
(Fellow, IEEE) received the B.Eng. degree (First Class Hons.) and the Ph.D. degree from the University of Glasgow, U.K., in 1994 and 1997, respectively. He is currently a Chair Professor of the Department of Computing, The Hong Kong Polytechnic University. He was the Editor-in-Chief of IEEE Transactions on Evolutionary Computation, and currently serves on the Editorial Board member of 10+ journals. He is currently the Vice-President (Publications) of IEEE Computational Intelligence Society, an Honorary Professor at University of Nottingham in UK, and the Chief Co-Editor of Springer Book Series on Machine Learning: Foundations, Methodologies, and Applications. He is an IEEE Fellow.\end{IEEEbiography}

\vfill

\end{document}